\documentclass[10pt,journal,compsoc]{IEEEtran}

\usepackage[nocompress]{cite}
\usepackage[pdftex]{graphicx}
\usepackage{amsmath}
\usepackage[ruled]{algorithm2e}
\usepackage{array}
\usepackage[caption=false,font=footnotesize,labelfont=sf,textfont=sf]{subfig}
\usepackage{stfloats}
\usepackage{url}

\usepackage{amsfonts,bm}
\usepackage{enumitem}
\usepackage{multirow} 
\usepackage{booktabs}
\usepackage{textcomp}
\usepackage[numbers]{natbib}
\usepackage{balance}

\newcolumntype{I}{!{\vrule width 1.5pt}}
\newlength\savedwidth

\newlength\savewidth
\newcommand\shline{\noalign{\global\savewidth\arrayrulewidth
                            \global\arrayrulewidth 1.5pt}%
                   \hline
                   \noalign{\global\arrayrulewidth\savewidth}}

\begin{document}

\title{Fine-Grained Urban Flow Inference}
\author{Kun~Ouyang,
        Yuxuan~Liang,
        Ye~Liu,
        Zekun~Tong,
        Sijie~Ruan,
        Yu~Zheng,~\IEEEmembership{Senior Member,~IEEE}
        and~David~S.~Rosenblum,~\IEEEmembership{Fellow,~IEEE}
        \IEEEcompsocitemizethanks{\IEEEcompsocthanksitem This paper is an extended version of an earlier paper published at the 25th SIGKDD conference (KDD 2019)~\cite{liang2019urbanfm}
        \IEEEcompsocthanksitem Kun Ouyang, Yuxuan Liang, Ye Liu and David S. Rosenblum are with the School of Computing, National University of Singapore, Singapore 119077. E-mail:\{ouyangk,yuxliang,liuye,david\}@comp.nus.edu.sg
        \IEEEcompsocthanksitem Zekun Tong is with the Department of Industrial System Engineering, National University of Singapore, Singapore 119077. E-mail: vantong96@outlook.com
        \IEEEcompsocthanksitem Sijie Ruan and Yu Zheng are with JD Intelligent Cities Research \& JD Intelligent Cities Research, Beijing, China. E-mail:\{sijieruan,msyuzheng\}@outlook.com}

\thanks{Manuscript received December 27, 2019.}}

\markboth{IEEE Transaction on Knowledge and Data Engineering}
{Shell \MakeLowercase{\textit{et al.}}: Bare Demo of IEEEtran.cls for Computer Society Journals}
%

\IEEEtitleabstractindextext{%
\begin{abstract}
The ubiquitous deployment of monitoring devices in urban flow monitoring systems induces a significant cost for maintenance and operation. A technique is required to reduce the number of deployed devices, while preventing the degeneration of data accuracy and granularity. In this paper, we present an approach for inferring the real-time and fine-grained crowd flows throughout a city based on coarse-grained observations. This task exhibits two challenges: the spatial correlations between coarse- and fine-grained urban flows, and the complexities of external impacts. To tackle these issues, we develop a model entitled UrbanFM which consists of two major parts: 1) an inference network to generate fine-grained flow distributions from coarse-grained inputs that uses a feature extraction module and a novel distributional upsampling module; 2) a general fusion subnet to further boost the performance by considering the influence of different external factors. This structure provides outstanding effectiveness and efficiency for small scale upsampling. However, the single-pass upsampling used by UrbanFM is insufficient at higher upscaling rates. Therefore, we further present UrbanPy, a cascading model for progressive inference of fine-grained urban flows by decomposing the original tasks into multiple subtasks. Compared to UrbanFM, such an enhanced structure demonstrates favorable performance for larger-scale inference tasks.
\end{abstract}}


\maketitle

\IEEEraisesectionheading{\section{Introduction}\label{sec:introduction}}



\IEEEPARstart{F}{ine}-grained urban flow monitoring systems are a crucial component of the information infrastructure systems of smart cities, providing a foundation for urban planning and various others applications such as traffic management. To obtain data at a spatially fine level of granularity, a system requires large numbers of sensing devices to be deployed in order to cover a citywide landscape. For example, thousands of piezoelectric sensors and loop detectors are deployed on road segments in a city to monitor fine-grained vehicle traffic flow volumes in real time. With a large number of devices deployed, a high cost is incurred due to the long-term operation (e.g., electricity and communication cost) and maintenance (e.g., on-site maintenance and warranty). A recent study showed that in Anyang, Korea, the annual operation and device maintenance costs for their smart city projects reached 100K USD and 400K USD respectively in 2015~\cite{idb-korea}. With the rapid development of smart cities on a worldwide scale, the cost of manpower and energy will become a prohibitive factor for further smartening the Earth. To reduce such expense, people require a novel technology that allows reducing the number of \textit{deployed} sensors while, most importantly, keeping the original data granularity unchanged. Therefore, how to approximate the original fine-grained information from available coarse-grained data (obtained from fewer sensors) becomes an urgent problem. 

\begin{figure}[!t]
	\centering
	\includegraphics[width=0.49\textwidth]{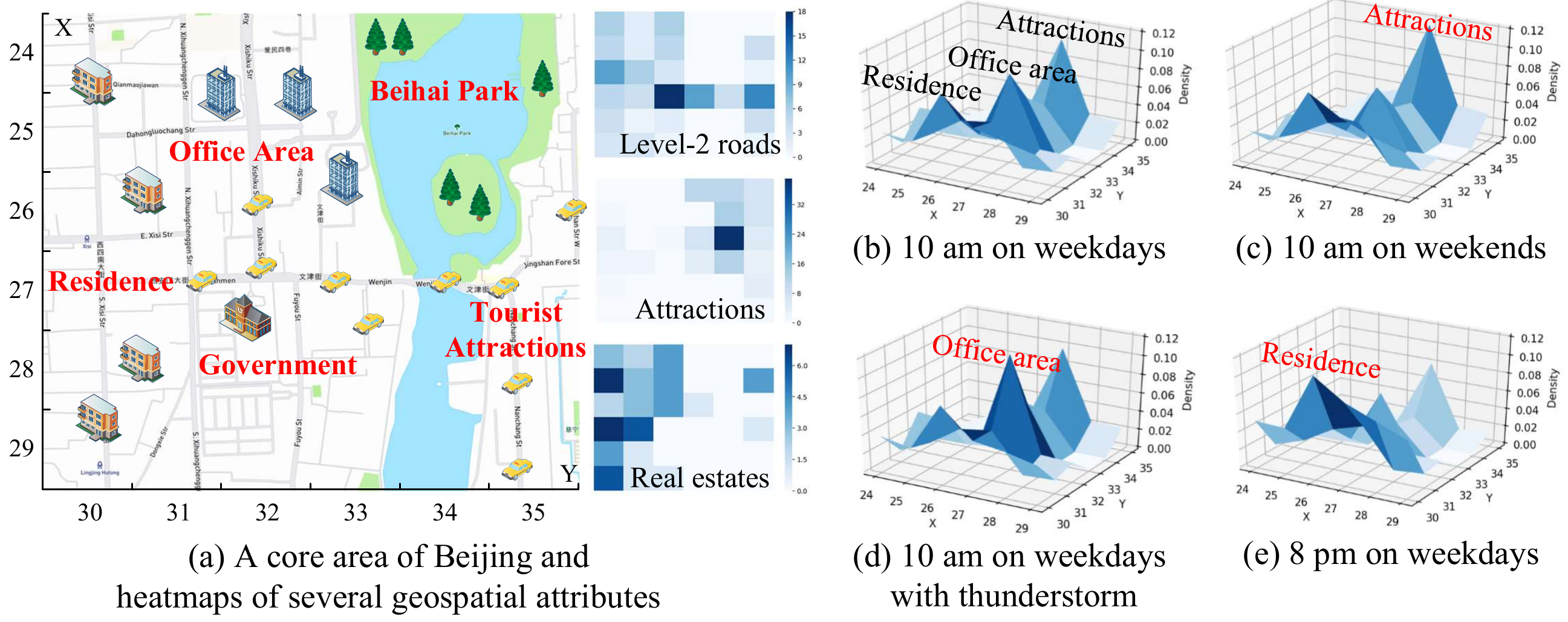}
	\vspace{-1.5em}
	\caption{\label{fig:ext_inf} Impacts of external factors on regional flow distributions. (a) We obtain Point of Interests (POIs) for different regions, and then categorize regions with different semantics according to the POI information. (b)-(e) depict the average flow distribution under various external conditions.}
	\vspace{-1em}
\end{figure}

Take monitoring traffic on a university campus as an example. We can reduce the number of interior loop detectors and keep sensors only at the entrances to save cost. However, we still desire to recover the fine-grained flow distribution within the campus given only the coarse-grained information. In this paper, \textbf{our goal} is to infer the real-time and spatially fine-grained flows from observed coarse-grained data on a citywide scale with many other regions (as shown in Figure \ref{fig:intro}). This Fine-grained Urban Flow Inference (FUFI) problem, however, is very challenging due to the following reasons:
\begin{figure}[!t]
	\centering
	\includegraphics[width=0.47\textwidth]{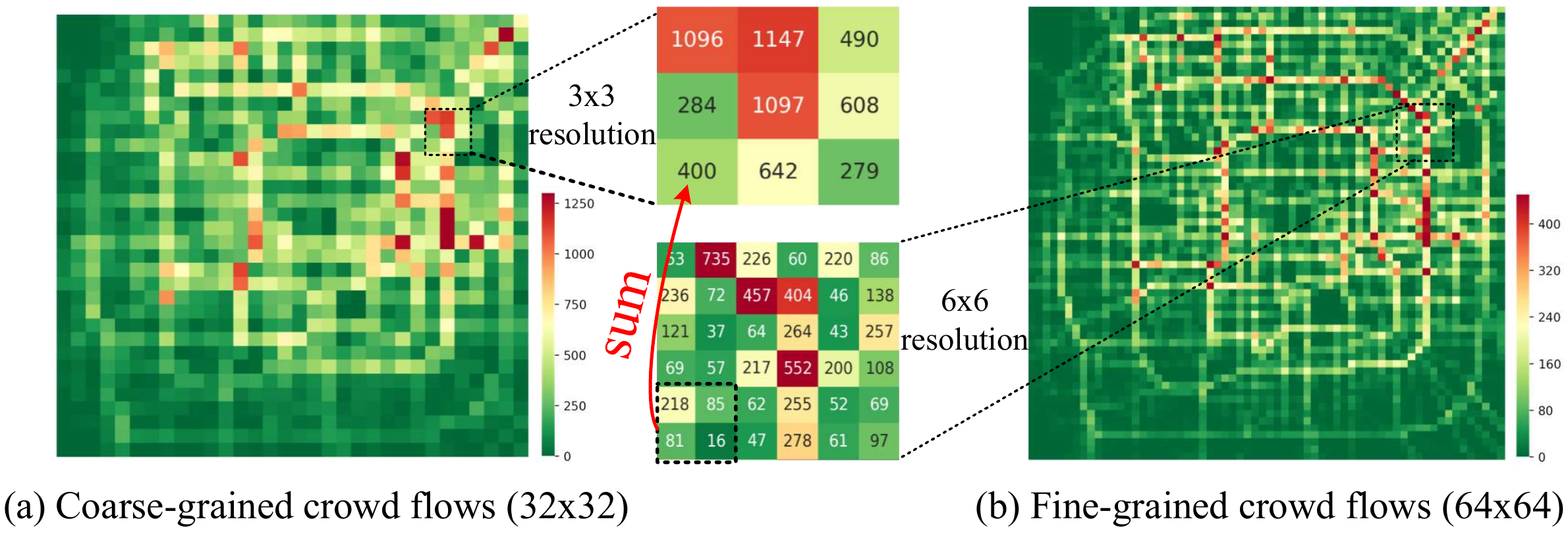}
	\vspace{-1em}
	\caption{\label{fig:intro} Traffic flows at two levels of granularities in Beijing, where each grid denotes a region.}	
	\vspace{-2em}
\end{figure}
\begin{itemize}[leftmargin=*]
\item{\textbf{Spatial Correlations.} Fine-grained flow maps have spatial and structural correlations with their coarse-grained counterparts. Essentially, the flow volume in a coarse-grained superregion (e.g., the campus), is distributed among constituent subregions (e.g., libraries, sports center) at the fine-grained level. This implies a crucial structural constraint (i.e., \textbf{spatial hierarchy} \cite{zheng2014urban}): the sum of the flow volumes among subregions strictly equals that of the corresponding superregion, as shown in Figure \ref{fig:intro}. Furthermore, the flow in one region can be affected by the flows in the nearby regions, which will impact the inference for the fine-grained flow distributions over subregions. Methods failing to capture these considerations would exhibit poor performance.}

\item{\textbf{External Factors.} The distribution of the flows in a given region is affected by various external factors, such as local weather, time of day and special events. To understand such impacts, we present a real-world study in an area of Beijing as shown in Figure~\ref{fig:ext_inf}(a). On weekdays, (b) shows more flows occuring at 10 a.m. in the office area and attractions as compared to at 8 p.m. when residences experience much higher flow density than the other areas (see (e)); on weekends, however, (c) depicts that people tend to be present in a park in the morning. All of these reflects our common sense that people go to work in the morning, to attractions for relaxation at the weekend, and return home at night. In addition, (d) shows that people are keen to move to indoor areas instead of the outdoor park during storms. These observations demonstrate that regions with different semantics present different flow distributions in the presence of different external factors. Moreover, these external factors can compound and thus influence the actual distribution in complicated ways.}
\end{itemize}

Inspired by techniques from the domain of image recovery, in the preliminary work we attack the FUFI problem by designing a neural network-based model entitled \textbf{Urban} \textbf{F}low \textbf{M}agnifier (UrbanFM)~\cite{liang2019urbanfm} which resolves the above challenges with an innovative network structure. Firstly, we extract features from  coarse-grained inputs using Convolutional Neural Networks (CNN) and perform upsampling based on high-level features. But in contrast to image processing, where the direct output is the target fine-grained image, we instead change the learning objective to inference of the \textit{distributions} of the fine-grained flows that capture how the flows in each superregion are distributed to their corresponding subregions. To this end, we present a \textit{distributional upsampling} module with a novel and parameter-free layer entitled $N^2$-\textit{Normalization} which provides superior performance over the image super-resolution baselines by exploiting the underlying structure of the FUFI problem. Moreover, we employ an external factor fusion subnet to capture the complexity of external impacts and produce a feature map that embeds the different impacts on different locations. Benefitting from the dedicated network architecture, UrbanFM outperforms all six baselines we have studied including heuristics and state-of-the-art methods across all three evaluation metrics we have considered.. 

Nevertheless, one limitation of our preliminary work is that it was evaluated only for 4x upsampling. When the required upsampling scale becomes larger (e.g., 8x), UrbanFM can encounter difficulties as it performs upsampling in a \textit{single forward pass}. In other words, the upsampling is conducted by consecutively stacking multiple sub-pixel upsampling layers. Such a simple strategy complicates the tasks of  feature extraction for frontal layers when the upsampled space becomes much larger. Inspired by the concept of Pyramid structure~\cite{lai2017deep}, in this paper, we present an enhanced model named \textbf{Urban} \textbf{Py}ramid Network (UrbanPy) which inherits key advantages from UrbanFM while performing progressive upsampling instead of single-pass. This model employs multiple key innovations to address the following deficiencies of its predecessor. 

\begin{itemize}[leftmargin=*]
\item \textbf{Pyramid Architecture.} In contrast to UrbanFM, UrbanPy decompose the overall upsampling objective into multiple subprocesses. For objective decomposition, UrbanPy employs a pyramid architecture consisting of multiple components, where each component functions as an atomic upsampler for a small scale (e.g. 2x). Such decomposition allows the network to divide a difficult task into much easier subtasks such that each component can solve its own subtask more effectively. Each component consists of two subnets and processes the upsampling task through a propose-and-correct paradigm, where the proposal network aims to propose a prototype based on the previous output and the correct network learns to correct the prototype. Following the spirit of UrbanFM, both the proposal and correction subnets focus on modeling the distributions over corresponding subregions in the upsampled map.

\item \textbf{Local Structure.} UrbanFM utilizes classic convolution layers for feature extraction, where the kernel weights are shared globally without considering the local characteristics of each superregion in which the individual distribution applies. However, human flows can be highly correlated to the geographic nature of the location. For example, a park tends to have smaller flow volume compared to the street next to it. Extracting the region-specific features can be difficult when kernel weights are shared. To this end, we embed geographic knowledge (e.g., road network, POIs) to each location and employ a non-shared convolution layer. This results in each superregion enjoying a customized submodel for flow inference while tailoring to the local geographic structure.

\item \textbf{Distributional Loss.} Though UrbanFM explicitly implants the spatial hierarchy into its model architecture, the training loss used is mean square error (MSE). This can introduce inconsistency between the distributional nature that the network is expected to capture and the training objective. To bridge this gap, we explicitly train the model according to discrepancies in the distribution space. In particular, we compare the inferred distribution with the ground truth distribution using KL-divergence in \textit{every local superregion}. This helps to preserve the local nature at the loss function compared to the simple MSE function.
\end{itemize} 

Our key contributions are summarized as follows.
\begin{itemize}
\item We formalize the problem of Fine-grained Urban Flow Inference, which is critical for modern urban information infrastructure construction. We show that the essence of this problem is to uncover the distributions of super-regions over their associative sub-regions. 
\item We present UrbanFM, which exploits the problem structure and shows superior performance versus the baseline methods. Moreover, we identify the limitation of this preliminary work for large scale upsampling and present the improved method UrbanPy, which incorporates multiple key innovations over UrbanFM\@.
\item We conducted extensive experiments using real-world datasets, including city-scale (i.e., Beijing) and district-scale (i.e. HappyValley, a theme park). Empirical results demonstrate the superiority of our methods compared to multiple state-of-the-art approaches.
\end{itemize}

\noindent\textbf{Outline.} We first formalize the FUFI problem in Section~\ref{sec:form}. Then we present in Section~\ref{sec:urbanfm} our preliminary work (i.e. UrbanFM) that addresses the problem by resolving the two challenges mentioned above. Furthermore, we discuss the limitations of UrbanFM and present the detail techniques employed by UrbanPy in Section~\ref{sec:urbanpy}. Extensive experiments are presented in Section~\ref{sec:experiment} to demonstrate the effectiveness of our method, followed by discussions of related works in Section~\ref{sec:relatedwork}. Section~\ref{sec:conclusion} concludes the paper.

\section{Problem Formulation}\label{sec:form}
This section first defines some notation and then formulates the problem of Fine-grained Urban Flow Inference (FUFI).

\noindent\textbf{Definition 1 (Region)} As shown in Figure~\ref{fig:intro}, we partition an area of interest (e.g., a city) evenly into an $I\times J$ grid map based on longitude and latitude, where a grid ele.lment denotes a region~\cite{zhang2017deep}. Partitioning the city into smaller regions (i.e., using larger $I,J$) allows ones to obtain flow data with more detail, which produces a more fine-grained flow map.

\noindent\textbf{Definition 2 (Flow Map)} Let $\mathbf{X}\in\mathbb{R}^{I\times J}_{+}$ represent a flow map of a particular time, where each entry $x_{i,j}\in \mathbb{R}_{+}$ denotes the flow volume of the flow agents (e.g., vehicle, people, etc.) in region $(i,j)$.

\noindent\textbf{Definition 3 (Superregion \& Subregion)} In our FUFI problem, a coarse-grained grid map indicates the data granularity  we can observe upon sensor reduction. It is obtained by combining nearby grids within an $N$-by-$N$ range of a fine-grained grid map using a scaling factor $N$. Figure~\ref{fig:intro} illustrates an example when $N=2$. Each coarse-grained grid in Figure~\ref{fig:intro}(a) is composed of $2\times 2$ smaller grids from Figure~\ref{fig:intro}(b). We define the aggregated larger grid as a \textit{superregion}, and its constituent smaller regions as \textit{subregions}. Note that in this setting, superregions do not share subregions. Hence, the relationship between superregions and the corresponding subregions induces a special \textit{structural constraint} in FUFI.

\noindent\textbf{Definition 4 (Structural Constraint)} The flow volume $x^c_{i,j}$ in a superregion of the coarse-grained grid map and the flows $x^f_{i',j'}$ in the corresponding subregions of the fine-grained counterpart obey the following equation:
\begin{equation}
	\label{eqn:constrain}
	x^c_{i,j} = \sum_{i',j'}{x^f_{i',j'}} \quad s.t. \lfloor{\frac{i'}{N}}\rfloor=i,\lfloor{\frac{j'}{N}}\rfloor=j.
\end{equation}
For simplicity, $i=1,2,\dots,I$ and $j=1,2,\dots,J$ in our paper unless otherwise specified.

\noindent\textbf{Problem Statement (Fine-grained Urban Flow Inference)}

\noindent Given an upscaling factor $N\in Z_+$ and a coarse-grained flow map $\mathbf{X}^c\in\mathbb{R}^{I\times J}_{+}$, infer the fine-grained counterpart $\mathbf{X}^f\in\mathbb{R}^{NI\times NJ}_{+}$ as accurately as possible subject to the structural constraints.

\section{Urban Flow Magnifier}\label{sec:urbanfm}
\begin{figure*}[!t]
	\centering
	\includegraphics[width=\textwidth]{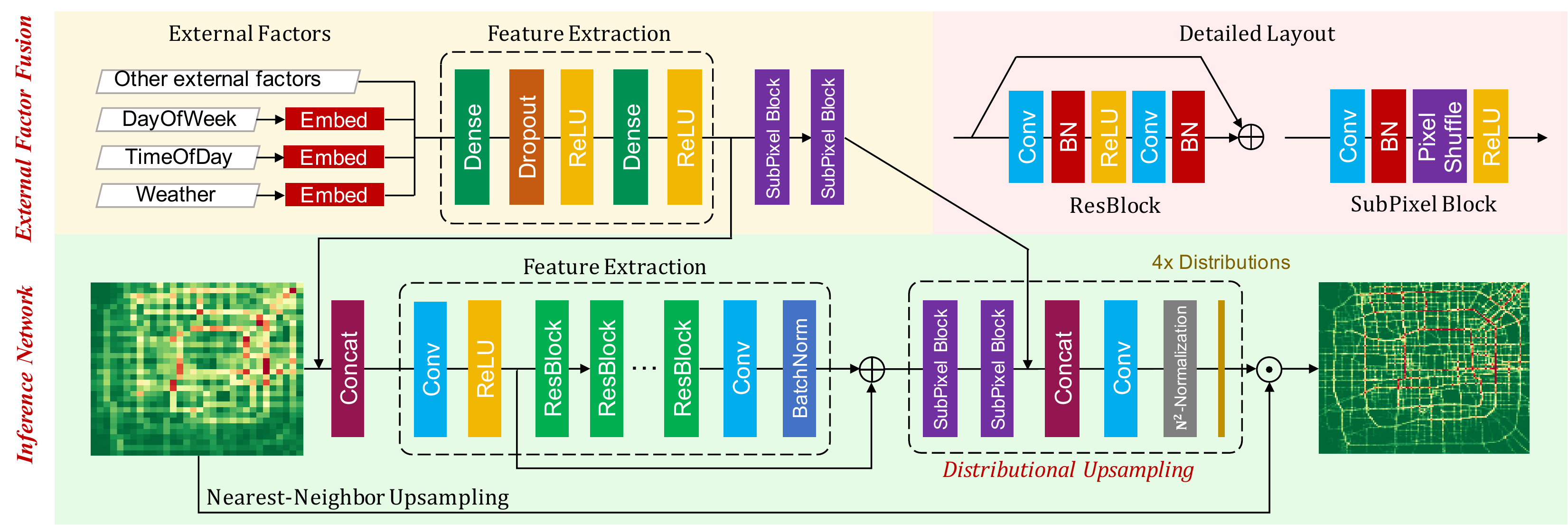}
	\vspace{-1em}
	\caption{\label{fig:urbanfm}The UrbanFM framework for $4\times$ upscaling ($N=4$). $\oplus$ denotes addition and $\odot$ denotes Hadamard product. Note that our framework allows other integer upscaling factor, not limited to power of 2.}
	\vspace{-1em}
\end{figure*}
Figure~\ref{fig:urbanfm} depicts the framework of UrbanFM which consists of two main components for conducting structurally constrained inference of fine-grained flows and capturing complex external influence on the flows, respectively. The inference network takes the coarse-grained flow map $\mathbf{X}^c$ as input, and then extract high-level features across the whole area by leveraging deep residual networks~\cite{he2016deep}. Taking extracted features as a priori knowledge, the \textit{distributional upsampling} module outputs a flow distribution over the subregions of each superregion by introducing a dedicated $N^2$-\textit{Normalization} layer. Finally, the Hadamard product of the inferred distribution with the original coarse-grained flow map gives the fine-grained flow map $\tilde{\mathbf{X}}^f$ as the network output. In an external factor fusion branch, we leverage embeddings and a dense network to extract pixel-wise external features at both coarse and fine granularity. The integration of external and flow features enables UrbanFM to exhibit fine-grained flow inference more effectively. In this section, we describe the designs of the two components, as well as the optimization scheme used in network training.

\vspace{-1mm}
\subsection{Inference Network\label{sec:infnet}}
The inference network aims to produce a map of fine-grained flow distributions over subregions from a coarse-grained input. We follow the general procedure in image super resolution (SR) methods, which is composed of two phases: 1) feature extraction; 2) inference upon upsampled features.

\vspace{-1mm}
\subsubsection{Feature Extraction}
In the input stage, we use a convolutional layer (with $9\times 9$ filter size and filter size $F$) to extract low-level features from the given coarse-grained flow map $\mathbf{X}^c$, and perform the first stage of fusion if external features are provided. Then $M$ residual blocks with identical layout take the (fused) low-level feature maps as input and construct high-level feature maps. The residual block layout, as shown on the top right of Figure~\ref{fig:urbanfm}, follows the guideline in \citeauthor{ledig2017srgan}~\cite{ledig2017srgan} which contains two convolutional layers ($3 \times 3$, $F$) followed by a batch normalization layer~\cite{ioffe2015batch}, with an intermediate ReLU~\cite{hahnloser2000digital} function to introduce non-linearity.

Since we utilize a fully convolutional architecture, the reception field grows larger as we stack the network deeper. In other words, each pixel at the high-level feature map will be able to capture distant or even citywide dependencies. Moreover, we use another convolutional layer ($3\times 3$, $F$) followed by batch normalization to guarantee non-trivial feature extraction. Finally, drawing from the intuition that the output flow distribution exhibits region-to-region dependencies on the original $\mathbf{X}^c$, we employ a skip connection to introduce identity mapping~\cite{he2016identity} between the low-level features and high-level features, building an information highway skipping over the residual blocks to allow efficient gradient back-propagation.

\vspace{-1mm}
\subsubsection{Distributional Upsampling}
In the second phase, the extracted features first go through $n$ sub-pixel blocks to perform an $N=2^n$ upscaling operation which produces a hidden feature $\mathbf{H}^f \in \mathbb{R}^{F\times~NI\times~NJ}$. The sub-pixel block, as illustrated in Figure~\ref{fig:urbanfm}, leverages a convolutional layer ($3\times 3$, $F\times2^2$) followed by batch normalization to extract features. Then it uses a PixelShuffle layer~\cite{shi2016espcn} to rearrange and upsample the feature maps to $2\times$ size and applies a ReLU activation at the end. After processing each sub-pixel block, the output feature maps are 2 times larger spatially with the number of channels unchanged. A convolutional layer ($9\times 9$, $F_o$) is applied post-upsampling, which maps $\mathbf{H}^f$ to a tensor $\mathbf{H}^f_o \in \mathbb{R}^{F_o\times~NI\times~NJ}$. $F_o=1$ in our case for simplicity. In SR tasks, $\mathbf{H}^f_o$ is usually the final output for the recovered image with super-resolution. However, the structural constraint that is essential to FUFI has not yet been considered.
%
%

In order to impose the structural constraint on the network, one straightforward manner is to add a \emph{structural loss} $L_s$ as a regularization term to the loss function:
%
\begin{equation}
	L_s = \sum_{i,j}{\bigg\lVert x^c_{i,j} - \sum_{i',j'}{\tilde{x}^f_{i',j'}}\bigg\rVert_F} \quad s.t. \lfloor{\frac{i'}{N}}\rfloor=i,\lfloor{\frac{j'}{N}}\rfloor=j.
\end{equation}

However, simply applying $L_s$ does not improve the model performance, as we demonstrate in Section~\ref{sec:experiment}. Instead, we design an $N^2$-\textit{Normalization} layer, which outputs a \textit{distributions} over every patch of $N$-by-$N$ subregions of an associated superregion. To achieve this, we reformulate Equation~\ref{eqn:constrain} as in the following:

\begin{equation}
	\begin{aligned}
	x^c_{i,j} = \sum_{i',j'}{\alpha^{\;}_{i',j'}x^c_{i,j}}\\
	s.t. \sum{\alpha_{i',j'}}=1, \;\alpha\in~\mathbb{R}_{+}, \lfloor{\frac{i'}{N}}\rfloor&=i,\lfloor{\frac{j'}{N}}\rfloor=j.
	\end{aligned}
\end{equation}
The flow volume in each subregion is now expressed as a \textit{fraction} of that in the superregion, i.e., $x^f_{i',j'}=\alpha^{\;}_{i'j'}x^c_{i,j}$, and we can treat the fraction as a probability. This allows us to interpret the network output in a meaningful way: the value in each subregion pixel states how likely the overall superregion flow will be allocated to the subregion $(i',j')$. With this reformulation, we shift our focus from directly generating the fine-grained flow to generating the flow distributions $\mathbf{D}'$. This essentially changes the network learning target and thus diverges from the traditional SR literature. To this end, we present the $N^2$-\textit{Normalization} layer: $N^2$-\textit{Normalization}$(\mathbf{H_o^f}) = \mathbf{D}'$, such that
\begin{equation}
	\mathbf{D}'_{(i,j)}=\mathbf{H}^{f}_{o,(i,j)}/\sum_{i'=(\lfloor{i/N}\rfloor-1)*N+1, \atop j'=(\lfloor{j/N}\rfloor-1)*N+1}^{i'=\lfloor{i/N}\rfloor*N, \atop j'=\lfloor{j/N}\rfloor*N}{\mathbf{H}^{f}_{o,(i',j')}}
	\label{eqn:n2}
\end{equation}

The $N^2$-Normalization layer induces no extra parameters to the network. Moreover, it can be easily implemented within a few lines of code (see Algorithm 1). Also, the operations can be fully paralleled and automatically differentiated at runtime. Remarkably, this reformulation relieves the network from concerning varying output scales and enables it to focus on producing a probability within $[0,1]$ constraint.
\begin{algorithm}[!t]
\SetAlgoLined
 \KwIn{x, scale\_factor, $\epsilon$}
 \KwOut{out} 
 \tcp{x: an input feature map} \vspace{-0.5mm}
 \tcp{scale\_factor: the upscaling factor} \vspace{-0.5mm}
 \tcp{$\epsilon$: a small number for numerical stability} \vspace{-0.5mm}
 \tcp{out: the structural distributions}
 \vspace{2mm}
 sum = SumPooling(x, scale\_factor)\;
 sum = NearestNeighborUpsampling(sum, scale\_factor)\;
 out \hspace{1mm}= x $\oslash$ (sum+$\epsilon$) // \texttt{element wise division} 
 \caption{$N^2$-Normalization}
 \label{alg:n2}
\end{algorithm}

Finally, we upscale $\mathbf{X}^c$ using nearest-neighbor upsampling~\cite{wikipedia_2019} ($\mathtt{NN\_Upsample}$) with scaling factor $N$ as the initial interpolation, and then generate the fine-grained inference by 
\begin{equation}
\tilde{\mathbf{X}}^f=\mathtt{NN\_Upsample}(\mathbf{X}^c; N)\odot\mathbf{D}'.
\end{equation}

\subsection{External Factor Fusion}
External factors, such as weather, can have a complicated and vital influence on the flow distribution over the subregions. For instance, even if the total population in a city remains stable over time, during stormy weather people tend to move from outdoor regions to indoor regions. When different external factors entangle, the actual impact on the flow becomes more difficult to capture. Therefore, we design a subnet to handle external factors \textit{all at once}.

In particular, we first separate the available external factors into two groups, continuous features and categorical features. Continuous features including temperature and wind speed are directly concatenated to form a vector $\mathbf{e}_{con}$. As shown in Figure \ref{fig:urbanfm}, categorical features include the day of week, the time of the day and kind of weather (e.g, sunny, rainy). Inspired by previous studies \cite{liang2018geoman}, we transform the categorical attributes into low-dimensional vectors by feeding them into seperate embedding layers, and then concatenate those embeddings to construct the categorical vector $\mathbf{e}_{cat}$. Then, the concatenation of $\mathbf{e}_{con}$ and $\mathbf{e}_{cat}$ gives the final external embedding, with $\mathbf{e}=[\mathbf{e}_{con};\mathbf{e}_{cat}]$.

Once we obtain the concatenation vector $\mathbf{e}$, we feed it into a feature extraction module whose structure is depicted in Figure~\ref{fig:urbanfm}. By using dense layers, the different external factors are compounded to construct a hidden representation, which models their complicated interaction. The module provides two outputs: the coarse-grained feature maps $\mathbf{H}^c_e$ and the fine-grained feature maps $\mathbf{H}^f_e$, where $\mathbf{H}^f_e$ is obtained by passing $\mathbf{X}^c_e$ through $n$ sub-pixel blocks similar to the ones in the inference network. Intuitively, $\mathbf{H}^c_e$ (respectively $\mathbf{H}^f_e$) is the spatial encoding for $\mathbf{e}$ in the coarse-grained (fine-grained) setting, modeling how each superregion (subregion) individually responds to the external factors. Therefore we concatenate $\mathbf{H}^c_e$ with $\mathbf{X}^c$, and $\mathbf{H}^f_e$ with $\mathbf{H}^f$ in the inference network. The early fusion of $\mathbf{H}^c_e$ and $\mathbf{X}^c$ allows the network to learn to extract a high-level feature describing not only the citywide flow, but also the external factors. In addition, the fine-grained $\mathbf{H}^f_e$ carries the external information all the way to the rear of the inference network, playing a similar role as an information highway, and thus prevents information perishing in the deep network.

\subsection{Loss Function}
UrbanFM provides an end-to-end mapping from coarse-grained input to fine-grained output, which is differentiable everywhere. Therefore, we can train the network through auto back-propagation, by providing training pairs ($\mathbf{X}^c, \mathbf{X}^f$) and calculating empirical loss between ($\mathbf{X}^f,\tilde{\mathbf{X}}^f$), where $\mathbf{X}^f$ is the ground truth and $\tilde{\mathbf{X}}^f$ is the outcome inferred by our network. As pixel-wise Mean Square Error (MSE) is a widely used cost function in many SR tasks, we employ the same in this work as follows:
\begin{equation}
L_{MSE}(\mathbf{X}^f,\mathbf{\tilde{X}}^f;\bm{\Theta})=\lVert{\mathbf{X}^f-\tilde{\mathbf{X}}^f}\rVert_F^2
\end{equation}
where $\bm{\Theta}$ denotes the set of parameters in UrbanFM.

\section{Urban Pyramid Network}\label{sec:urbanpy}
\begin{figure*}[!t]
	\centering
	\includegraphics[width=\textwidth]{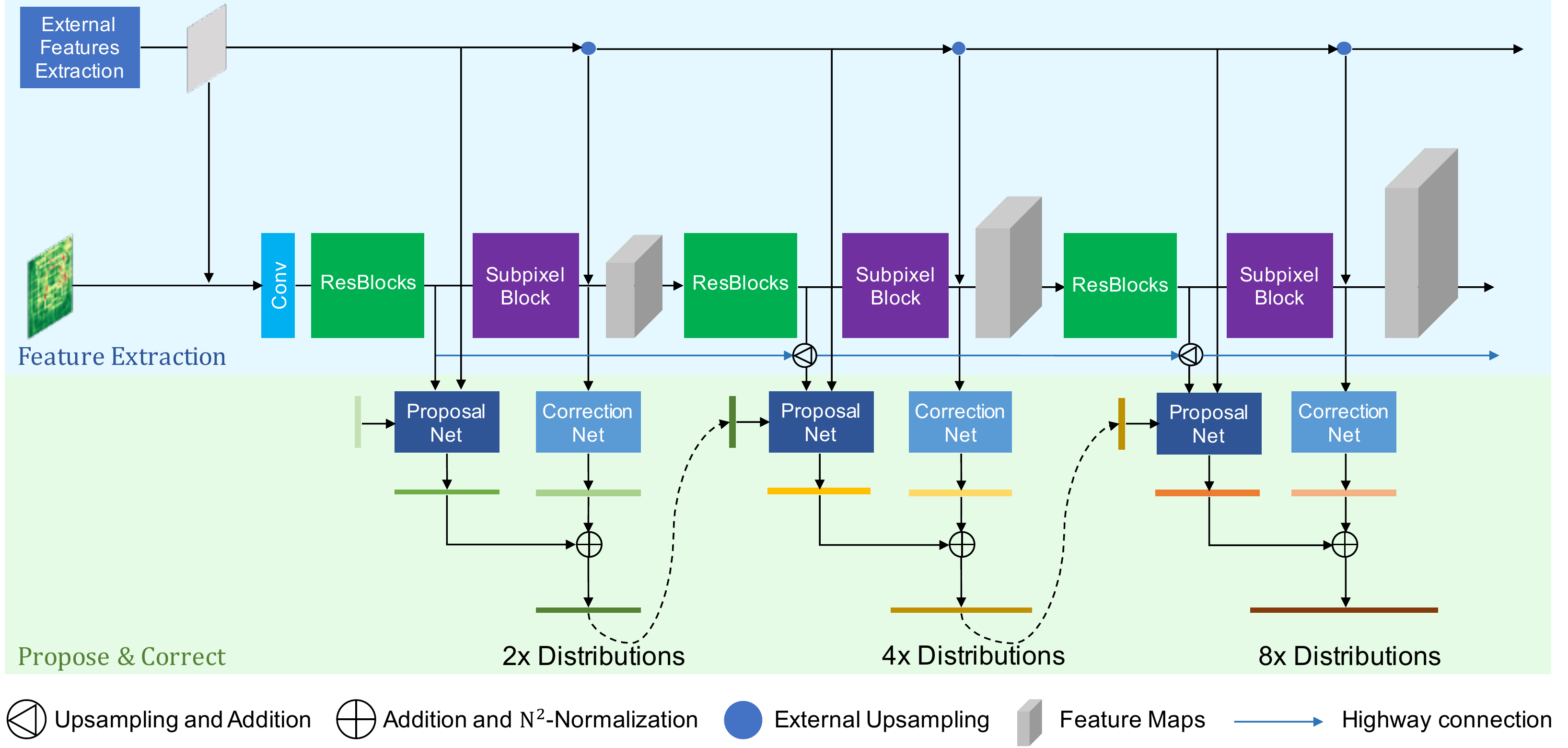}
	\vspace{-2em}
	\caption{\label{fig:urbanpy} A UrbanPy framework for $2\times$, $4\times$ and $8\times$ upscaling. We employ a cascading strategy to progressively upsample the coarse-grained inputs. At each level, we first prepare features from the input flow map and external factors. Then a propose-and-correct component is used to cooperatively produce two views of the target distribution map. We aggregate and renormalize the two views using $N^2$-Normalization to give the map of mixture distributions which will then be an input for the next level. The foremost distribution map $\mathbf{D}_0$ is initialized with an all-ones matrix. Note that we omit nearest neighbor upsampling and the Hadamard product to avoid redundant presentation of elements shown in Figure~\ref{fig:urbanfm}.}
	\vspace{-1em}
\end{figure*}
The design of UrbanFM has followed two key principles: first, reconstruct a fine-grained flow map according to high-level features; second, embed the structural constraint in the model design. Maintaining those two principles, we now present UrbanPy which advances the UrbanFM framework by resolving three limitations: 1) a single upsampling process, 2) non-distinguishable features and 3) inconsistent MSE loss. The overall architecture is depicted in Figure~\ref{fig:urbanpy}.

\subsection{Pyramid Architecture}
Taking a coarse-grained flow map as input, our model decomposes the high-scale upsampling task into $L$ consecutive upsampling subtasks with upsampling factors $[s_1, s_2,\ldots, s_l,\ldots s_L]$ respectively, where $s_L=N$.\fnbelowfloat
\footnote{For simplicity, upsample rates for the width and height are described as being the same here but need not be in the general case.} In accordance with the decomposition, UrbanPy employs $L$ components of similar structure where each component upsamples from $s_l$ to $s_{l+1}$. The common component structure includes two modules: 1) feature extraction and 2) propose-and-correct. The feature extraction module abstracts a higher-level representation of the original input and transforms the feature maps from a lower scale to a higher scale. Then the processed representations are fed into a proposal network and a correction network, which collaboratively infer the flow distributions at the $l^{th}$ scale. We describe the two modules further in the following.

\subsubsection{Feature Extraction}
The input to the $l^{th}$ feature extraction component a feature tensor $\textbf{H}_{l-1}\in \mathcal{R}^{s_{l-1}I\times s_{l-1}J\times F_{l-1}}$ generated from the previous feature extraction component. Given $\textbf{H}$, we first uses $M_2$ number of residual blocks of the same layout to construct a high-level representation $\tilde{\textbf{H}}_l\in \mathcal{R}^{s_{l-1}I\times s_{l-1}J\times F_{l}}$ without changing the spatial dimensions. In order to meet the upscaled dimension at the current level, we employ a Subpixel block to enlarge the spatial dimension of $\tilde{\textbf{H}}_l$ and produce upsampled feature maps $\textbf{H}_l\in\mathcal{R}^{s_{l}I\times s_{l}J\times F_{l}}$. $\tilde{\textbf{H}}_l$ and $\textbf{H}_l$ thus provide two different views of the inputs, and are used as separate features for the following proposal network and correction network respectively. For simplicity, we set $F_l=F$ for $l=1,\ldots,L.$

\textbf{Highway Connection.} Larger upscaling rates require stacking more feature extraction component, leading to a deeper network architecture. Although we have utilized residual blocks to facilitate gradient passage during back-propagation, the existence of other layers (e.g., upsampling) can affect the dynamics such that the deeper network becomes harder to train. Therefore, we add highway connections (denoted as blue arrows in Figure~\ref{fig:urbanpy}) across components such that previous features can be directly reused in the deeper layers. Specifically, given a list of previous representations $[\mathbf{\tilde{H}}_1,\ldots,\mathbf{\tilde{H}}_{l-1}]$, we upsample them to the scale of $s_l$ and then aggregate with the current representation $\mathbf{\tilde{H}}_l$ by addition, which gives $\mathbf{\tilde{H}}^*_l$. This process resembles a moving average process with equal decaying factors.

\textbf{External Factor Fusion.} We employ the same external factor fusion module used in UrbanFM for extracting the external features. Let $\mathbf{H}^1_e$ be $\mathbf{H}^c_e$. At each level we recruit a subpixel block to upsample $\mathbf{H}_e^{l-1}$ to $\mathbf{H}^l_e$. We use the same weights in each upsampler to reduce the number of parameter as we observed no obvious advantage of using distinct weights in our experiments.

\subsubsection{Propose and Correct} 
\begin{figure}[!b]
	\centering
	\includegraphics[width=0.48\textwidth]{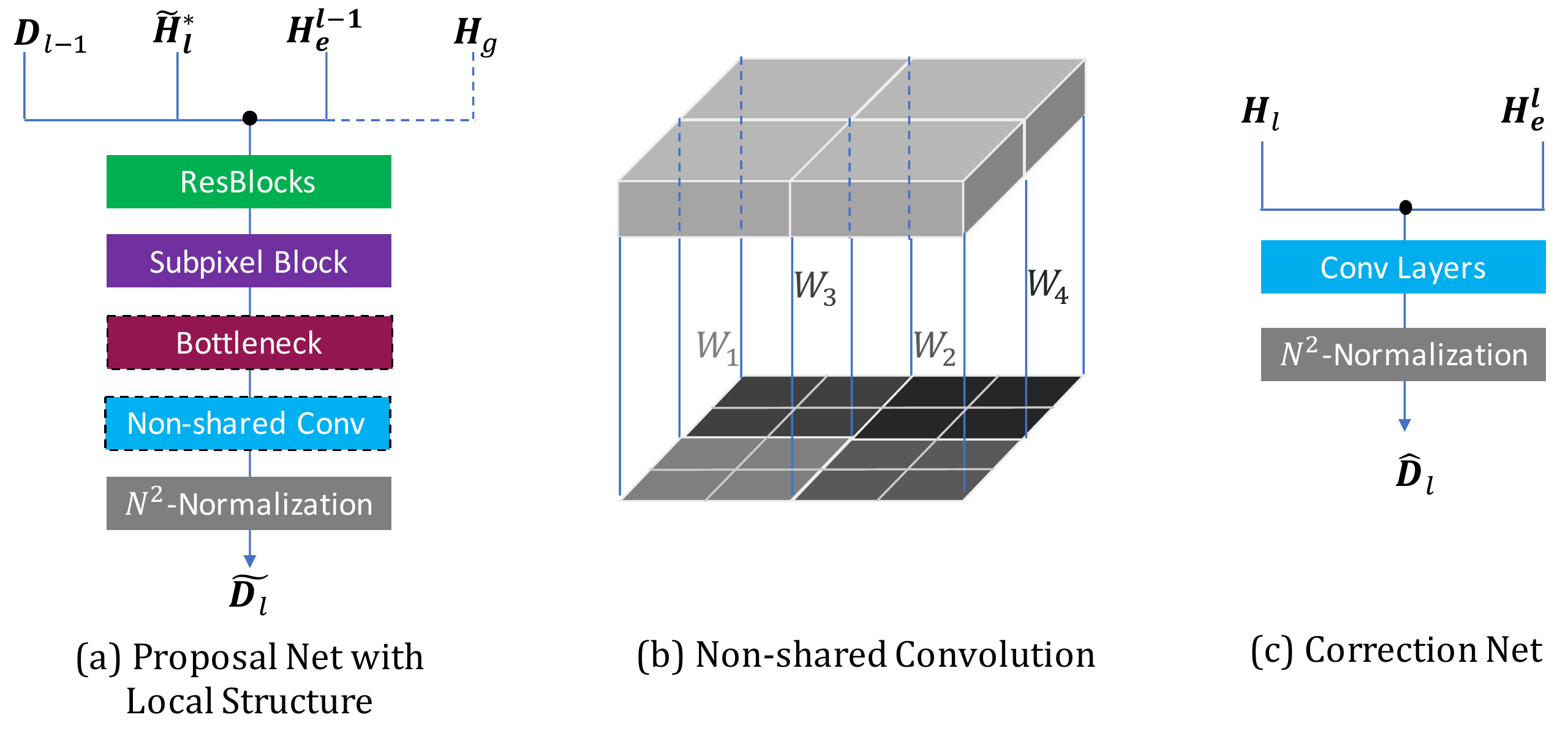}
	\vspace{-1em} 
	\caption{\label{fig:proposalnet} Structure of the correction net, the proposal net and the non-shared convolution layer. (a) and (c) uses black dots to denote concatenation; $\mathbf{H}_g$ is the extra geographic feature for obtaining local structure. (b) uses different color of $W$ to denotes different kernel weights; similar to urbanFM, it outputs a feature map of channel size $F_{o}$ before $\mathrm{N^2}$-Normalization.}
	\vspace{-1em}
\end{figure}
Inspired by the idea of a Laplacian pyramid, where the difference between a blurred image and its original image is modeled, we apply a similar idea as a propose-and-correct architecture such that the proposal network aims to generate a prototype distribution map, and the correction network is responsible for correcting the prototype. 

Specifically, given representations $\tilde{\textbf{H}}^*_l$ and $\mathbf{H}_e^{l-1}$, the proposal network produces a prototype $\tilde{\textbf{D}}_{l}\in \mathcal{R}^{s_{l}I\times s_{l}J}$ by embracing the flow distributions $\mathbf{D}_{l-1}\in \mathcal{R}^{s_{l-1}I\times s_{l-1}J}$ produced at the previous level (see the dotted lines in Figure~\ref{fig:urbanpy}). Note that these connections give the whole network a cascade structure, which improves the consistency between different levels as there are obvious correlations between flows at different granularities. The interior structure of the proposal network is depicted in Figure~\ref{fig:proposalnet}(a), where we stack $R$ number of residual blocks to strengthen the capacity of the proposal network, followed by an $\mathrm{N}^2$-Normalization layer to enforces the hierarchical structure of the output distributions at the current scale. The dotted elements are described in the Section~\ref{sec:localstructure}.

Taking $\textbf{H}_l$ and $\mathbf{H}_e^l$ as input, the correction network employs a similar structure as the proposal network (i.e. convolutional layers followed by $\mathrm{N}^2$-Normalization) to generate a correction distribution map $\textbf{D}'_{l}\in \mathcal{R}^{s_{l}I\times s_{l}J}$. But the correction network is designed to be light-weight. One reason is that adjusting the distribution based on the proposal network is a simpler task. Moreover, the correction network actually can take advantage of the weights of the upsample block from the feature extraction branch whose output also serves as direct input to the correction network. Therefore, we design the correction network with just one convolution layer to transform the upsampled feature maps. 

Once we obtain the prototype $\tilde{\mathbf{D}}_{l}$ and correction $\mathbf{\hat{D}}_{l}$, we add these two distribution maps in an element-wise manner, and then renormalize the results to generate the final distribution map $\textbf{D}_{l}$. That is,
\begin{equation}
\mathbf{D}'_{l} = \mathtt{N^2}{\text -}\mathtt{Normalization}(\tilde{\mathbf{D}}_{l} + \mathbf{\hat{D}}_{l}).
\end{equation}
Akin to UrbanFM, the generated distribution map is Hadamard-multiplied by the interpolated coarse-grained input to give the inferred fine-grained flow prediction at the $s_l\times$ scale.

\subsubsection{A Probabilistic View}\label{sec:probview} 
Note that the UrbanPy architecture is different from Laplacian pyramid models~\cite{lai2018fast,lai2017deep,wang2018fully}. The latter \emph{add} the interpolated image with the predicted residual image to obtain the final inference. However, in the problem of FUFI we concerned with modeling flow distributions, but distributions do not exhibit \emph{closure} under addition (i.e., $p_1 + p_2 \notin \mathcal{P}$ where $p_1, p_2\in \mathcal{P}$ and $\mathcal{P}$ is the set of distributions). Instead, UrbanPy infers the final distribution map as a symmetric mixture density of the prototype and the correction, where each super-region is represented by a mixture probability distribution.

\subsection{Local Structure}\label{sec:localstructure}

Each super-region in the coarse-grain map can cover a very large area. For instance, each grid in the 8-by-8 granularity contains about 6.25 $km^2$ area in Beijing. The geographic properties (building layout, road plan) of a grid can vary significantly from one grid to another. To capture such specialty, we include the geographic features as additional knowledge and employ a non-shared convolution layer to allow customized feature extraction for each super-region. 

\subsubsection{Geographic Embeddings}
For each level $s_l$, we obtain level-specific geographic features including POI and road network. For POI, we obtain a set of POI density maps of the city across different categories (e.g., education and entertainment), which results in (raw) feature maps $\mathbf{H}_{poi}\in\mathcal{R}^{s_lI\times s_lJ\times C_{poi}}$, where $C_{poi}$ denotes the total number of POI categories. Likewise, in terms of the road network structure, we obtain the tier-1, tier-2 and tier-3 road density for each region and construct a road network feature tensor $\mathbf{H}_{rn}\in\mathcal{R}^{s_lI\times s_lJ\times 3}$. Then the geographic feature $\mathbf{\tilde{H}}_g = [\mathbf{H}_{poi};\mathbf{H}_{rn}]$ is given by concatenation of both along the last dimension. $\mathbf{\tilde{H}}_g$, however, is very sparse. To mitigate the sparsity, we perform feature compression by pre-training a Stacked Denoising Auto-encoder~\cite{vincent2010stacked} and then use the hidden code $\mathbf{H}_g=\mathtt{encoder}(\mathbf{\tilde{H}}_g)\in\mathcal{R}^{s_lI\times s_lJ\times C_g}$ as a compressed knowledge of the raw features. Eventually, each grid can be represented by an individual embedding $\mathbf{h}^g_{i,j}\in\mathbf{H}_g$ of size $N_g$. 

\subsubsection{Non-shared Convolution}
The UrbanFM model uses the classic convolution layers such that the kernel weights are shared globally, which can be insufficient to capture the peculiarity of each superregion. Therefore, besides recruiting geographic features, we use separated weights to produce a more local representation of each super-region before applying $\mathrm{N^2}$-Normalization, as is shown in Figure~\ref{fig:proposalnet}. 

First, we define the classic discrete convolution operation $*$ according to the formulation from ~\cite{yu2015multi} as follows\footnote{To keep consistent, we use even kernel width here. An odd kernel width simply modifies $\Omega_r=[-r,r]^2\cap\mathbb{Z}^2$.}. Let $F:\mathbb{Z}^2\rightarrow \mathcal{R}$ be a discrete function. Let $\Omega_r=[1-r, r]^2\cap \mathbb{Z}^2$ and let $k:\Omega_r\rightarrow \mathcal{R}$ be a discrete filter of size $(2r)^2$. Then the $*$ is defined as: 
\begin{equation}
(F*k)(\mathbf{p})=\sum_{\mathbf{s}+\mathbf{t}=\mathbf{p}}{F(\mathbf{s})k(\mathbf{t})}.
\end{equation}
To generalize the classic convolution, we introduce a set of kernels $\mathcal{K}=\{k_{i,j}\}$ where $(i,j)\in\mathbb{Z}^2$ and define the local non-shared convolution $*'$ as:
\begin{equation}
(F*'\mathcal{K})(\mathbf{p})=\sum_{\mathbf{s}+\mathbf{t}=\mathbf{p},\atop \mathbf{p}=(r+2ri,r+2rj)}{F(\mathbf{s})k_{i,j}(\mathbf{t})}.
\end{equation}
The kernel width $2r$ is thus equal to the convolution stride. Specifically, in each level $l$, we set $2r$=$s_l$ to force each kernel to focus on a specific super-region as shown in Figure~\ref{fig:proposalnet}(c). To reduce the parameter cost for customization, a bottleneck layer~\cite{he2016deep} is deployed in advance to compress the channel of the feature maps provided by previous layers to 2 for simplicity. 

\subsection{Distributional Loss}
UrbanFM measures the MSE between the ground truth flow and the predicted \emph{flow values}, and uses it as the loss to optimize the network parameters. Such loss is straight forward, but ignores the underlying structure of the problem. To resolve such inconsistency, we step deeper into the problem by directly measure the divergence existing between the truth distribution and the predicted \emph{flow distributions}. 

In particular, in level $l$, we first prepare the ground truth $\mathbf{D}_l$=$\mathbf{X}_l/\mathtt{NN\_Upsample}(\mathbf{X}_1; s_l)$, where $\mathtt{NN\_Upsample}(\cdot;s_l)$ indicates nearest neighbour upsampling by scale $s_l$. Once we obtain the inferred distribution map $\mathbf{D}'_l=[\mathbf{d}'_1,\ldots,\mathbf{d}'_{I\times J}]$ that contains $I\times J$ different distributions, the total distributional loss $L_D$ between the two set of distributions is computed via KL-divergence:
\begin{align*}
L^l_{D}(\mathbf{D}_l', \mathbf{D}_l;\bm{\Theta}) &= \sum_{i,j=1,1}^{I,J}{KL(\mathbf{d}'_{i,j}, \mathbf{d}_{i,j})}\\
where\quad KL(\mathbf{d}',\mathbf{d})			&= {\sum_{p=1}^{s_l^2}d'_{p}\log\frac{d'_p}{d_p}}.
\end{align*}
The distributional loss exploits the essence of the model and is defined on each superregion-subregion distilling. Using $L_D$ alone can train the network till convergence, but its asymmetry can produce unstable gradients which slows down the training process~\cite{bishop2006pattern}. Therefore, we combine both the MSE loss and distributional loss across at each level, and aggregate through all levels to constitute the overall loss function:
\begin{equation}
\textstyle L=\sum_l{\alpha L_D^l+(1-\alpha)L_{MSE}^l},
\end{equation}
where $\alpha$ is the coefficient to control the scale of the two losses. In experiments, we set $\alpha$=1e-$2$ by default to balance the magnitude of the gradients from the two losses, giving stable multi-task training according to~\cite{chen2018gradnorm}.

\section{Experiments}\label{sec:experiment}
Our experiments aim to quantitively and qualitatively examine the capacity of the presented two models in a citywide scenario. Therefore, we conduct extensive experiments using taxi flows in Beijing to comprehensively test the model from different aspects. Different from the preliminary evaluations that focus only on $4\times$ upsampling, we conduct experiments involving four different scales. In addition to citywide, we conduct further experiments in a theme park, namely Happy Valley, to show the adaptivity of our models in a relatively small area.

\subsection{Experimental Settings}
\subsubsection{Datasets}
\begin{figure}[h!]
	\centering
	\includegraphics[width=0.45\textwidth]{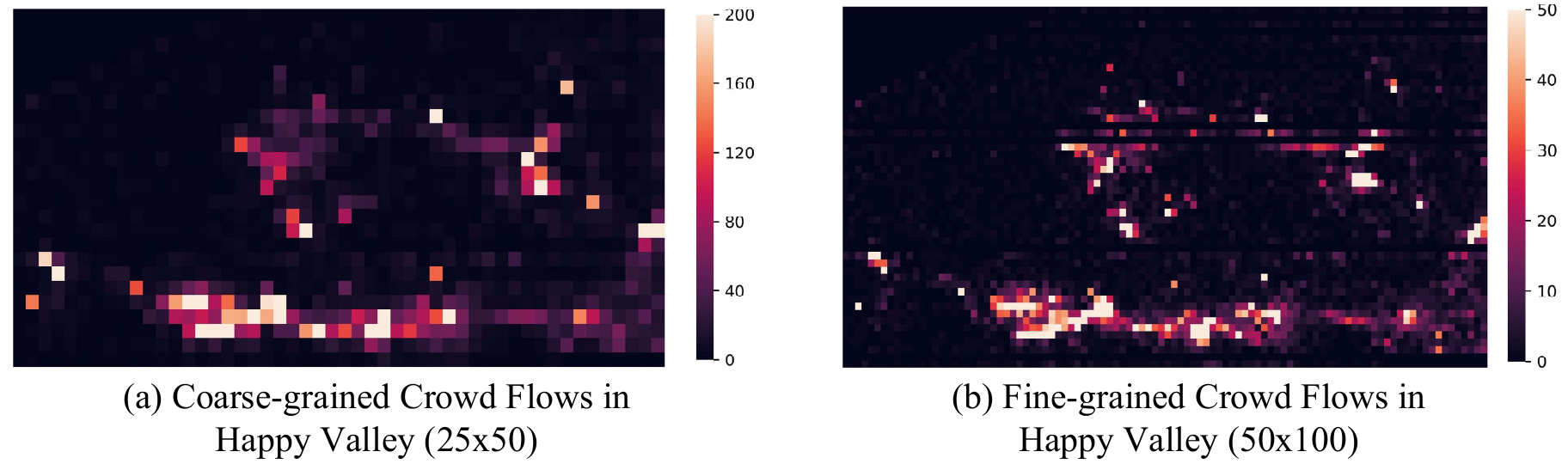}
	\caption{Visualization of crowd flows in HappyValley.}
	\vspace{-1em}	
	\label{fig:dataset}
\end{figure}

Table \ref{tab:dataset} details the two datasets we use, namely TaxiBj and HappyValley, where each dataset contains two sub-datasets: urban flows and external factors. Since a number of fine-grained flow data are available as ground truth, in this paper, we can obtain the a coarse-grained flows by aggregating subregion flows from the fine-grained counterparts. As our empirical evaluations span across multiple scales, we obtain data at each granularity separately. When conducting experiments for $s_l\times$ upscaling, we aggregate the subregions in a $s_l\times s_l$ area to generate the flows of the corresponding superregion.
\begin{itemize}[leftmargin=*]
	\item \textbf{TaxiBJ}\footnote{See our GitHub https://github.com/yoshall/UrbanFM}: This dataset indicates the taxi flows traveling throughout Beijing. Figure~\ref{fig:intro} gives an example when the studied area is split into 32$\times$32 grids. where each grid reports the coarse-grained flow information every 30 minutes within four different periods: P1 to P4 (detailed in Table \ref{tab:dataset}). In our experiments, we partition the data into non-overlapping training, validation and test data by a ratio of 2:1:1 respectively for each period. For example, in P1 (7/1/2013-10/31/2013), we use the first two-month data as the training set, the next month as the validation set, and the last month as the test set. With this dataset, we construct coarse-grained data of 4 different granularities (i.e., $8\times 8$, $16\times 16$, $32\times 32$ and $64\times 64$) as the coarse-grained inputs, targeting upsampling factor $N=16,8,4,2$ respectively. The data partition for each granularity is the same.
	\item \textbf{HappyValley}: We obtain this dataset by crawling from an open website\footnote{heat.qq.com} which provides hourly gridded crowd flow observations for a theme park named Beijing Happy Valley, with a total 5$\times10^5 m^2$ area coverage, from 1/1/2018 to 10/31/2018. As shown in Figure~\ref{fig:dataset}, we partition this area with 25$\times$50 uniform grids in coarse-grained setting, and target a fine granularity at 50$\times$100 with an upscaling factor $N=2$. Note that in this dataset, one special external factor is the ticket price, including day price and night price, obtained from the official account of HappyValley in WeChat. Regarding the smaller area, crowd flows exhibit large variance across samples given the 1-hour sampling rate. Thus, we use a ratio of 8:1:1 to split training, validation, and test set to provide more training data.
\end{itemize}
\vspace{-1em}
\begin{table}[!h]
	\centering
	\caption{Dataset Description.}
	\vspace{-1em}
	\tabcolsep=1.5mm
	  \begin{tabular}{lll}
	  \shline
	  \textbf{Dataset} & \textbf{TaxiBJ} & \textbf{HappyValley} \\
	  \hline
	  \multirow{4}[1]{*}{Time span} & P1: 7/1/2013-10/31/2013 &  \\
			& P2: 2/1/2014-6/30/2014 & 1/1/2018- \\
			& P3: 3/1/2015-6/30/2015 & 10/31/2018 \\
			& P4: 11/1/2015-3/31/2016 &  \\
	  \midrule
	  Time interval & 30 minutes & 1 hour \\
	  Coarse-grained size & 32$\times$32 & 25$\times$50 \\
	  Fine-grained size & 128$\times$128 & 50$\times$100 \\
	  Upscaling factor ($N$) & 4     & 2 \\
	  \midrule
	  \multicolumn{3}{l}{\textbf{External factors (meteorology, time and event)}} \\
	  Weather (e.g., Sunny) & 16 types & 8 types \\
	  Temperature/\textcelsius & [-24.6, 41.0] & [-15.0, 39.0] \\
	  Wind speed/mph & [0, 48.6] & [0.1, 15.5] \\
	  \# Holidays & 41    & 33 \\
	  Ticket prize/\textyen  & /     & [29.9, 260] \\
	  \midrule
	  \multicolumn{3}{l}{\textbf{Geographic features}}\\
	  Road network & density & /\\
	  Point of Interest & density & /\\
	  \shline
	  \end{tabular}%
	\label{tab:dataset}%
\end{table}%

\begin{table*}[htbp]
  \renewcommand{\arraystretch}{1.1}
  \centering
  \caption{\textbf{Qualititative results for model comparison.} We conducted inference experiments for four different scales, where all target the same endpoint with 128$\times$128 resolution. For each scale, the results for single-process and progressive upscaling are presented separately. Across all methods, we use \underline{\textbf{underlined bold}}, \textbf{bold} and \underline{underline} to indicate the best, the second best  and the third performance, respectively. This helps us to identify the performance change along the enlarging of upscaling.}
    \begin{tabular}{rr|ccc|ccc|ccc|ccc}
    \hline
    \hline
    \multicolumn{1}{r}{\multirow{2}[2]{*}{Methods}} & \multicolumn{1}{c|}{\multirow{2}[2]{*}{Upscales}} & \multicolumn{3}{c|}{P1} & \multicolumn{3}{c|}{P2} & \multicolumn{3}{c|}{P3} & \multicolumn{3}{c}{P4} \\
          &       & \multicolumn{1}{l}{RMSE} & \multicolumn{1}{l}{MAE} & \multicolumn{1}{l|}{MAPE} & RMSE  & MAE   & MAPE  & RMSE  & MAE   & MAPE  & RMSE  & MAE   & MAPE \\
    \hline
    \rule{0pt}{10pt}            
    MEAN  & 2     & 16.899 & 8.931 & 2.935 & 21.557 & 11.373 & 3.477 & 22.111 & 11.876 & 3.633 & 15.369 & 8.218 & 2.766 \\
    HA    & 2     & 3.494 & \underline{1.723} & \underline{0.306} & 3.932 & 1.933 & \underline{0.305} & 4.072 & 2.002 & \underline{0.299} & 3.063 & 1.547 & \underline{0.287} \\
    SRCNN & 2     & 3.216 & 1.793 & 0.433 & 3.500 & 1.998 & 0.468 & 3.587 & 2.034 & 0.446 & 2.861 & 1.643 & 0.422 \\
    VDSR  & 2     & 3.203 & 1.750 & 0.387 & 3.523 & 1.894 & 0.325 & 3.575 & 1.965 & 0.369 & 2.776 & 1.533 & 0.337 \\
    ESPCN & 2     & 3.170 & 1.789 & 0.433 & 3.472 & 1.969 & 0.432 & 3.564 & 2.014 & 0.419 & 2.813 & 1.608 & 0.398 \\
    SRResNet & 2  & \underline{3.101} & 1.742 & 0.430 & \underline{3.383} & \underline{1.840} & 0.351 & \textbf{3.481} & \underline{1.889} & 0.336 & \textbf{2.732} & \underline{1.518} & 0.347 \\
    UrbanFM & 2   & \underline{\textbf{3.015}} & \underline{\textbf{1.553}} & \underline{\textbf{0.265}} & \underline{\textbf{3.344}} & \underline{\textbf{1.729}} & \underline{\textbf{0.260}} & \underline{\textbf{3.415}} & \underline{\textbf{1.783}} & \underline{\textbf{0.262}} & \underline{\textbf{2.675}} & \underline{\textbf{1.397}} & \underline{\textbf{0.248}} \\
    \hline
    DeepSD & 2     & 3.216 & 1.793 & 0.433 & 3.500 & 1.998 & 0.468 & 3.587 & 2.034 & 0.446 & 2.861 & 1.643 & 0.422 \\
    LapSRN & 2     & 3.202 & 1.763 & 0.396 & 3.468 & 1.900 & 0.370 & \underline{3.584} & 1.959 & 0.360 & 2.797 & 1.545 & 0.338 \\
    UrbanPy & 2     & \textbf{3.093} & \textbf{1.578} & \textbf{0.268} & \textbf{3.420} & \textbf{1.758} & \textbf{0.264} & \underline{3.584} & \textbf{1.843} & \textbf{0.268} & \underline{2.759} & \textbf{1.428} & \textbf{0.252} \\
    \shline
    \rule{0pt}{10pt}    
    MEAN  & 4     & 20.918 & 12.019 & 4.469 & 20.918 & 12.019 & 5.364 & 27.442 & 16.029 & 5.612 & 19.049 & 11.070 & 4.192 \\
    HA    & 4     & 4.741 & 2.251 & 0.336 & 5.381 & 2.551 & 0.334 & 5.594 & 2.674 & \underline{0.328} & 4.125 & 2.023 & 0.323 \\
    SRCNN & 4     & 4.297 & 2.491 & 0.714 & 4.612 & 2.681 & 0.689 & 4.815 & 2.829 & 0.727 & 3.838 & 2.289 & 0.665 \\
    VDSR  & 4     & 4.159 & 2.213 & 0.467 & 4.586 & 2.498 & 0.486 & 4.730 & 2.548 & 0.461 & 3.654 & 1.978 & 0.411 \\
    ESPCN & 4     & 4.206 & 2.497 & 0.732 & 4.569 & 2.727 & 0.732 & 4.744 & 2.862 & 0.773 & 3.728 & 2.228 & 0.711 \\
    SRResNet & 4     & 4.164 & 2.457 & 0.713 & 4.524 & 2.660 & 0.688 & 4.690 & 2.775 & 0.717 & 3.667 & 2.189 & 0.637 \\
    UrbanFM & 4     & \textbf{3.991} & \textbf{2.036} & \textbf{0.331} & \underline{4.374} & \underline{2.256} & \underline{\textbf{0.322}} & \textbf{4.539} & \underline{2.348} & \underline{\textbf{0.323}} & \textbf{3.526} & \textbf{1.831} & \underline{\textbf{0.310}} \\
    \hline
    DeepSD & 4     & 4.156 & 2.368 & 0.614 & 4.554 & 2.612 & 0.621 & 4.692 & 2.739 & 0.682 & 3.877 & 2.297 & 0.652 \\
    LapSRN & 4     & \underline{3.997} & \underline{2.040} & \underline{0.339} & \underline{\textbf{4.353}} & \textbf{2.235} & \underline{0.324} & \textbf{4.539} & \textbf{2.343} & 0.330 & \underline{3.531} & \underline{1.841} & \underline{0.315} \\
    UrbanPy & 4     & \underline{\textbf{3.949}} & \underline{\textbf{1.997}} & \underline{\textbf{0.330}} & \textbf{4.359} & \underline{\textbf{2.227}} & \textbf{0.323} & \underline{\textbf{4.519}} & \underline{\textbf{2.319}} & \textbf{0.326} & \underline{\textbf{3.514}} & \underline{\textbf{1.821}} & \textbf{0.314} \\
    \shline
    \rule{0pt}{10pt}
    MEAN  & 8     & 22.565 & 13.205 & 5.221 & 28.903 & 16.871 & 6.305 & 29.677 & 17.617 & 6.587 & 20.606 & 12.168 & 4.882 \\
    HA    & 8     & 5.629 & 2.682 & 0.442 & 6.429 & 3.058 & 0.443 & 6.717 & 3.211 & 0.431 & 4.959 & 2.433 & 0.423 \\
    SRCNN & 8     & 6.103 & 3.433 & 1.027 & 6.569 & 3.708 & 0.971 & 6.959 & 4.012 & 1.086 & 5.518 & 3.181 & 0.935 \\
    VDSR  & 8     & 5.178 & 2.681 & 0.580 & 5.482 & 2.821 & 0.489 & 5.878 & 3.069 & 0.543 & 4.623 & 2.416 & 0.481 \\
    ESPCN & 8     & 4.854 & 2.664 & 0.664 & 5.291 & 2.854 & 0.580 & 5.529 & 2.981 & 0.570 & 4.311 & 2.368 & 0.547 \\
    SRResNet & 8     & 4.783 & 2.554 & 0.579 & 5.215 & 2.807 & 0.572 & 5.492 & 2.935 & 0.551 & 4.298 & 2.297 & 0.487 \\
    UrbanFM & 8     & \textbf{4.748} & \underline{2.373} & \textbf{0.377} & \underline{5.224} & \underline{2.647} & \textbf{0.366} & \underline{5.488} & \underline{2.776} & \textbf{0.358} & \textbf{4.195} & \textbf{2.130} & \textbf{0.342} \\
    \hline
    DeepSD & 8     & 5.412 & 3.056 & 0.831 & 5.680 & 3.175 & 0.756 & 6.023 & 3.397 & 0.804 & 4.733 & 2.452 & 0.450 \\
    LapSRN & 8     & \underline{4.772} & \textbf{2.355} & \underline{0.401} & \textbf{5.197} & \textbf{2.578} & \underline{0.375} & \textbf{5.456} & \textbf{2.721} & \underline{0.383} & \underline{4.209} & \underline{2.141} & \underline{0.369} \\
    UrbanPy & 8     & \underline{\textbf{4.572}} & \underline{\textbf{2.237}} & \underline{\textbf{0.352}} & \underline{\textbf{5.003}} & \underline{\textbf{2.476}} & \underline{\textbf{0.339}} & \underline{\textbf{5.259}} & \underline{\textbf{2.610}} & \underline{\textbf{0.342}} & \underline{\textbf{4.071}} & \underline{\textbf{2.052}} & \underline{\textbf{0.336}} \\
    \shline
    \rule{0pt}{10pt}
    MEAN  & 16    & 23.157 & 13.622 & 5.540 & 29.695 & 17.390 & 6.722 & 30.521 & 18.165 & 7.035 & 21.193 & 12.554 & 5.191 \\
    HA    & 16    & 6.307 & 2.920 & 0.459 & 7.289 & 3.358 & 0.461 & 7.619 & 3.534 & 0.448 & 5.597 & 2.658 & 0.442 \\
    SRCNN & 16    & 9.987 & 5.699 & 1.779 & 9.198 & 5.148 & 1.647 & 10.912 & 6.226 & 1.822 & 8.181 & 4.618 & 1.490 \\
    VDSR  & 16    & 6.313 & 2.994 & 0.551 & 6.780 & 3.216 & 0.468 & 7.074 & 3.429 & 0.517 & 5.758 & 2.867 & 0.572 \\
    ESPCN & 16	  & 5.373 & 2.914 & 0.762 & 5.956 & 3.211 & 0.750 & 6.256 & 3.422 & 0.806 & 4.892 & 2.699 & 0.715 \\ 
    SRResNet & 16    & 5.381 & 2.672 & 0.504 & 5.959 & 3.056 & 0.594 & 6.295 & 3.189 & 0.559 & 5.066 & 2.596 & 0.551 \\
    UrbanFM & 16    & \underline{5.344} & \underline{2.573} & \underline{0.383} & \underline{5.968} & \underline{2.898} & \underline{0.371} & \underline{6.241} & \underline{3.033} & \underline{0.366} & \textbf{4.839} & \underline{2.362} & \underline{0.368} \\
    \hline
    DeepSD & 16    & 5.983 & 3.223 & 0.811 & 6.392 & 3.380 & 0.725 & 6.818 & 3.644 & 0.787 & 5.400 & 2.671 & 0.441 \\
    LapSRN & 16    & \textbf{5.244} & \textbf{2.449} & \textbf{0.357} & \textbf{5.820} & \textbf{2.729} & \underline{\textbf{0.339}} & \textbf{6.104} & \textbf{2.879} & \textbf{0.343} & \underline{4.860} & \textbf{2.321} & \textbf{\textbf{0.346}} \\
    UrbanPy & 16    & \underline{\textbf{5.204}} & \underline{\textbf{2.434}} & \underline{\textbf{0.357}} & \underline{\textbf{5.750}} & \underline{\textbf{2.719}} & \textbf{0.352} & \underline{\textbf{6.061}} & \underline{\textbf{2.868}} & \underline{\textbf{0.346}} & \underline{\textbf{4.778}} & \underline{\textbf{2.290}} & \textbf{0.350} \\
    \hline
    \hline
    \end{tabular}%
  \label{tab:table1}%
\end{table*}%

\subsubsection{Evaluation Metrics}

We use three common metrics for urban flow data to evaluate the model performance from different facets. Specifically, Root Mean Square Error (RMSE) is defined as:
\begin{equation*}
\small
	RMSE = \sqrt{\frac{1}{z}\sum_{s=1}^{z}{\bigg\lVert \mathbf{X}^f_s-\tilde{\mathbf{X}}^f_s} \bigg\rVert^2_F},
\end{equation*} 
where $z$ is the total number of samples, $\tilde{\mathbf{X}}^f_s$ is $s$-th the inferred value and $\mathbf{X}^f_s$ is corresponding ground truth. Mean Absolute Error (MAE) and Mean Absolute Percentage Error (MAPE) are defined as: $MAE = \frac{1}{z}\sum_{s=1}^{z}{\lVert{\mathbf{X}^f_s-\tilde{\mathbf{X}}^f_s}}\rVert_{1,1}$ and $MAPE = \frac{1}{z}\sum_{s=1}^{z}{\lVert{(\mathbf{X}^f_s-\tilde{\mathbf{X}}^f_s}) \oslash \mathbf{X}^f_s} \rVert_{1,1}$. In general, RMSE favors spiky distributions, while MAE and MAPE focus more on the smoothness of the outcome. Smaller metric scores indicate better model performances.

\subsubsection{Baselines}
We compare our proposed model with seven baselines that belong to the following three classes: (1) Heuristics, (2) Single-pass upsampling and (3) Progressive upsampling. The heuristic methods are designed based on intuition or empirical knowledge. In the single-pass category, we include four state-of-the-art methods for single image super-resolution, the domain from which we are inspired to design UrbanFM. In the progressive upsampling branch, we involve two methods with different progressive strategies: stacking and cascading, where one is the state of the art on statistical upsampling for climate data and the other is for image super-resolution. We detail them as follows:

\vspace{1mm}
\noindent\emph{Heuristic methods}:
\begin{itemize}[leftmargin=*]
  \item \textbf{Mean Partition (Mean)}: We evenly distribute the flow volume from each superregion in a coarse-grained flow map to the $N^2$ subregions, where $N$ is the upscaling factor.
  \item \textbf{Historical Average (HA)}: Similar to distributional upsampling, HA treats the value over each subregion a fraction of the value in the respective superregion, where the faction is computed by averaging all training data.
\end{itemize}

\vspace{1mm}
\noindent\emph{Single-pass methods}:
\begin{itemize}[leftmargin=*]
  \item \textbf{SRCNN} \cite{dong2016srcnn}: SRCNN presented the first successful introduction of convolutional neural networks (CNNs) into the SR problems. It consists of three layers: patch extraction, non-linear mapping, and reconstruction. Filters of spatial sizes $9 \times 9$, $5 \times 5$, and $5 \times 5$ were used respectively. The number of filters in the two convolutional layers is 64 and 32 respectively. In SRCNN, the low-resolution input is upscaled to the high-resolution space using a single filter (commonly bicubic interpolation) before reconstruction.
  \item \textbf{ESPCN} \cite{shi2016espcn}: Bicubic interpolation used in SRCNN is a special case of the deconvolutional layer. To overcome the low efficiency of such deconvolutional layer, Efficient Sub-Pixel Convolutional Neural Network (ESPCN) employs a sub-pixel convolutional layer aggregates the feature maps from LR space and builds the SR image in a single step.
  \item \textbf{VDSR} \cite{kim2016vdsr}: Since both SRCNN and ESPCN follow a three-stage architecture, they have several drawbacks such as slow convergence speed and limited representation ability. Inspired by the VGG-net, \citeauthor{kim2016vdsr} presents a Super-Resolution method using Very Deep neural networks with depth up to 20. This study suggests that a large depth is necessary for the task of SR.
  \item \textbf{SRResNet} \cite{ledig2017srgan}: SRResNet enhances VDSR by using the residual architecture presented by~\citeauthor{he2016deep}\cite{he2016deep}. The residual architecture allows one to stack a much larger number of network layers, which bases many benchmark methods in computer vision tasks.
\end{itemize}

\vspace{1mm}
\noindent\emph{Progressive methods}:
\begin{itemize}[leftmargin=*]
  \item \textbf{DeepSD}\footnote{As stacking of UrbanFMs gives similar or worse results over LapSRN at large scales, we show the results for DeepSD and LapSRN.}~\cite{vandal2017deepsd}: DeepSD is the state-of-the-art method on statistical upcaling (i.e., super-resolutioin)for meteorological data. It employs the stacking strategy by independently training multiple SRCNNs, each aims at downscaling for an intermediate level. It performs further upsampling by simply stacking up those pretrained SRCNNs. This method, however, is slow as it needs to perform interpolation first and then extracts features on the large-size feature maps. Another related art~\cite{zong2019deepdpm} also employs this same technique for a different task.
  \item \textbf{LapSRN} \cite{lai2017deep}: LapSRN is named due to the Laplacian pyramid structure. At each level, it predicts a residual image and then adds with the interpolated output from the previous level to construct the current prediction. The training of LapSRN employs a cascading strategy such that the whole model is trained end to end. 
\end{itemize}

\subsubsection{Variants}
We study the following variants of UrbanFM to evaluate the roles of different components.

\begin{itemize}[leftmargin=*]
	\item \textbf{UrbanFM-ne}: We simply remove the external factor fusion subnet from our method, which can help reveal the significance of this component.
	\item \textbf{UrbanFM-sl}: Upon removing the external subnet, we further replace the distributional upsampling module by using sub-pixel blocks and $L_s$ to consider the structural constraint in this variant. 
\end{itemize}
Study on the variants of UrbanPy involves choosing the different depth of the proposal network $R$, different filter depth $M_2$ and filter size $F$ for each feature extraction module at each level. We denote the variants by $M_2$-$F$-$R$ and omit the name as shown in Table~\ref{tab:param_study} for a more succinct presentation. 

\subsubsection{Training Details \& Hyperparameters}

Our model, as well as the baselines, are completely implemented by PyTorch with one TITAN V GPU. We leverage Adam~\cite{kingma2014adam}, an algorithm for stochastic gradient descent, to perform network training with learning rate $lr$=$1e$-$4$ and use batch size being 16 for the single-pass methods. We also apply a staircase-like schedule by halving the learning rate every 20 epochs, which allows smoother search near the convergence point. In the external subnet, there are 128 hidden units in the first dense layer with dropout rate 0.3, and $I \times J$ hidden units in the second dense layer. We embed DayOfWeek to $\mathbb{R}^2$, HourOfDay to $\mathbb{R}^3$ and weather condition to $\mathbb{R}^3$. Besides, for VDSR and SRResNet, we use the default settings in their paper. Since SRCNN, ESPCN performs poorly with default settings, we test different hyperparameters for them and finally use 384 as the number of filters in their two convolutional layers.

For progressive methods, as the training are typically much slower, we double the learning rate as well as the batch size. We stack pre-trained SRCNN with 384 filter size to construct DeepSD, and default hyper-parameters for LapSRN. For UrbanPy, we use $D$=8 for $2\times$ tasks to endow more power for the network and $D$=4 for other scales unless otherwise specified. We save the best-performed model according to the validation results and early-stop the training if the best model is not altered after 50 epochs.

%
\begin{table}[b]
  \centering
  \vspace{-2em}
  \renewcommand{\arraystretch}{1.1}
  \caption{\textbf{P-value of Wilcoxon signed-rank test.} }
    \begin{tabular}{lrrr}
    \shline
    Test Group & \multicolumn{1}{l}{RMSE} & \multicolumn{1}{l}{MAE} & \multicolumn{1}{l}{MAPE} \\
    \hline
    UrbanFM-SRResNet & 5.6e-4 & 2.2e-4 & 2.2e-4 \\
    UrbanPy-LapSRN & 4.0e-4 & 2.2e-4 & 4.5e-3   \\
    \hline
    UrbanPy-UrbanFM & 3.5e-2 & 6.1e-3 & 3.9e-2  \\
    \shline
    \end{tabular}%
  \label{tab:test}%
\end{table}%

\subsection{Results on TaxiBJ}
\subsubsection{Model Comparison}
\noindent This subsection compares the model effectiveness against the baselines. We report the results of UrbanFM with $M$-$F$ being 16-64 and UrbanPy with $M_2$-$F$-$R$ being 4-64-4 as our default settings. Further experiments on variants regarding different $M$-$F$ and $M_2$-$F$-$R$ will be discussed later. Table~\ref{tab:table1} illustrates the overall performances of all methods for the TaxiBJ dataset for tasks with $2\times$, $4\times$, $8\times$ and $16\times$ upscaling. Due to space limitation, the key tests of significance regarding the results of this table is shown at Table~\ref{tab:test}.

We summarize the table with several key observations.
\begin{enumerate}[leftmargin=*]
\item \textit{Single-pass.} By comparing UrbanFM with heuristic and single-pass baselines, it can be seen that UrbanFM consistently outperforms \emph{all} methods in all metrics in all 16 groups of experiments. Take the strongest baseline SRResNet for example. By averaging across all experiments, UrbanFM advances it by 2\%, 9\% and 37\% on RMSE, MAE, and MAPE respectively. Accordingly, the first row in Table~\ref{tab:test} validates the significance of this result. Though the backbone structure are similar, the advances of UrbanFM over SRResNet indeed underline the effectiveness of the proposed distributional upsampling component and the usefulness of the features extracted by the external factor fusion module.

\item \textit{Progressive.} In the category of progressive upsampling, it can be seen that the LapSRN is the stronger baseline, which shows the betterment of using cascading strategy in our task, as it allows a thoroughly trained network. Nevertheless, LapSRN is beaten by UrbanPy in almost all metrics across all experiments. Specifically, the average improvements shown by UrbanPy are 2\%, 3\%, and 10\% on the three metrics. This improvement comes not only from the spirits that are inherited from UrbanFM, but also the unique structure we design for UrbanPy.

\item \textit{Single-pass vs Progressive.} By comparing progressive methods against the single-pass methods, we can see that progressive networks generally demonstrate improvements at larger scales upsampling (typically at 4$\times$ and larger). For example, DeepSD versus the SRCNN counterparts and cascading methods versus all the single-pass baselines. In particular, UrbanPy outperform \emph{all} single-pass baselines in this regard. This can be attributed to that progressive method allows the model to conduct upsampling in a smoother way instead of abruptly enlarging the output scale by a large factor. It also worths noting that UrbanFM remains very competitive at $4\times$ upscaling compared to the progressive baselines. This emphasizes that the proposed $N^2$-\emph{Normalization} and the external factor fusion can provide significant enhancement even without smoothing the upsampling task. 

\end{enumerate}

\noindent\textbf{Test of Significance} 

\noindent The Wilcoxon test is an alternative for paired t-test when samples are from a non-normal distribution. Given two methods [A-B], we aim to test the alternative hypothesis: "The error produced by A is significantly \textbf{smaller} than that of B" across the 16 experiments settings (four scales for four periods) and then report the p-values of the null hypothesis at Table~\ref{tab:test}. The first and second rows are testing using our method and the best baseline in the respective category. The third row compares UrbanPy against UrbanFM. All tests are significant at level $\alpha=0.05$.

\begin{figure}[!b]
	\centering
	\vspace{-1em}
	\begin{subfloat}[16-64 setting]{
		\includegraphics[width=0.23\textwidth]{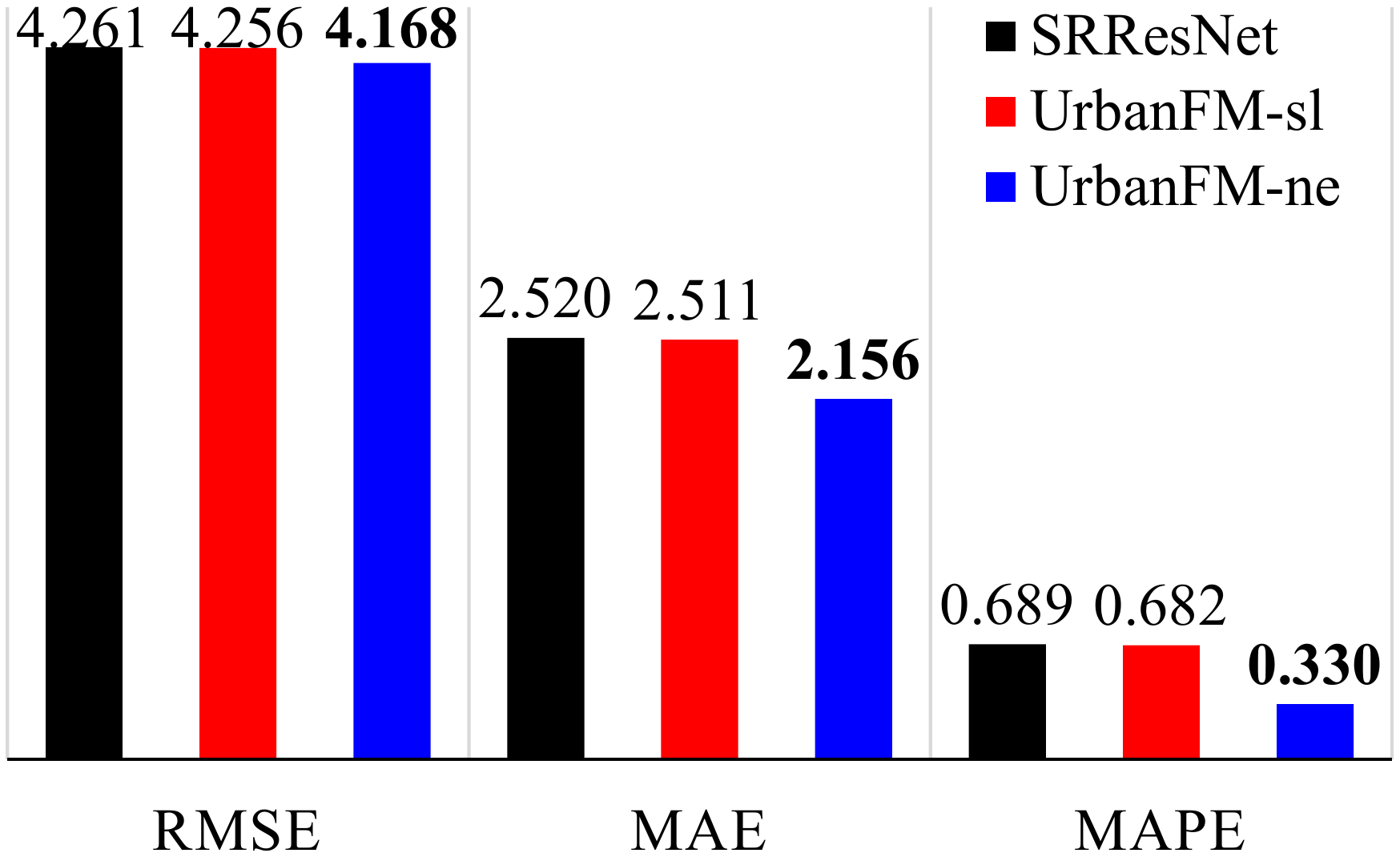}
		\label{fig:hiloss1}}
	\end{subfloat}
	\begin{subfloat}[16-128 setting]{
		\includegraphics[width=0.23\textwidth]{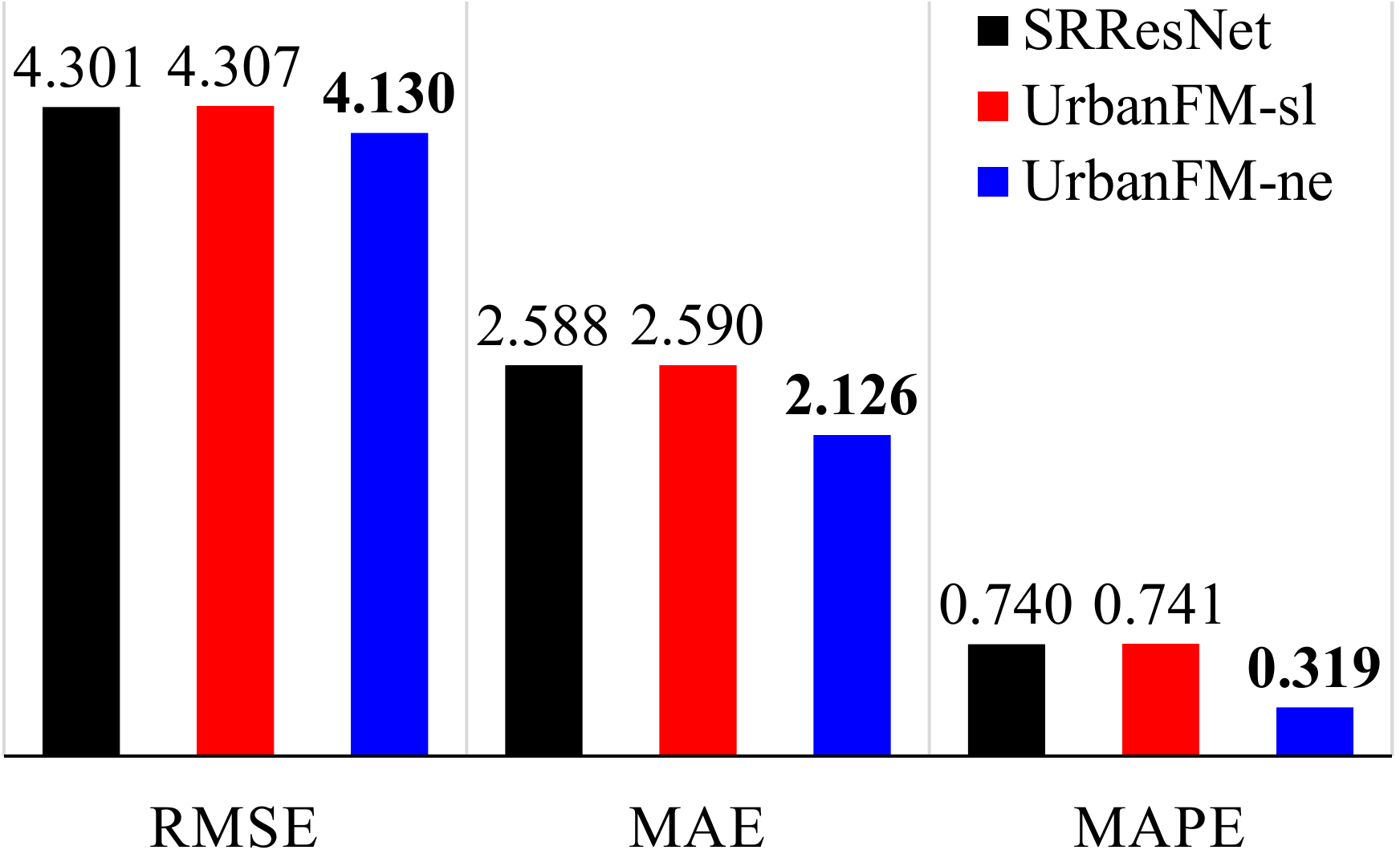}
		\label{fig:hiloss2}}
	\end{subfloat}
	\caption{\textbf{Study on Distributional Upsampling.} Performance comparison on whether or not applying the structural constraint.}
	\vspace{-1em}
	\label{fig:hiloss}
\end{figure}
\subsubsection{Studies on Effectiveness}
\noindent\textbf{Study on Distributional Upsampling}

\noindent To examine the effectiveness of the distributional upsampling module, we compare SRResNet with UrbanFM-ne (using distributional upsampling but no external factors) and UrbanFM-sl (using structural loss instead of distributional upsampling), as shown in Figure~\ref{fig:hiloss}. In both $M$-$F$ settings, it can be seen that UrbanFM-sl regularized by $L_s$ performs very close to the SRResNet which is not constrained at all. Though under the setting of 16-64, Urban-sl achieves a smaller error than SRResNet in a subtle way, under the 16-128 setting they behave the opposite. On the contrary, UrbanFM-ne consistently outperforms the others on all three metrics. 
This results has verified the superiority of the distributional upsampling module over $L_s$ for imposing the structural constraint.

\vspace{1mm}
\noindent\textbf{Study on External Factor Fusion}

\noindent External impacts, though are complicated, can assist the network for better inferences when they are properly modeled and integrated, especially in a more difficult situation when there is less data budget. Thereby, we study the effectiveness of external factors by randomly subsampling from the original training set according to four ratios (i.e., 10\%, 30\%, 50\% and 100\%) which corresponds to four difficulty levels: hard, semi-hard, medium and easy.

\begin{figure}[t!]
	\centering
	\begin{subfloat}[Results on RMSE]{
		\includegraphics[width=0.225\textwidth]{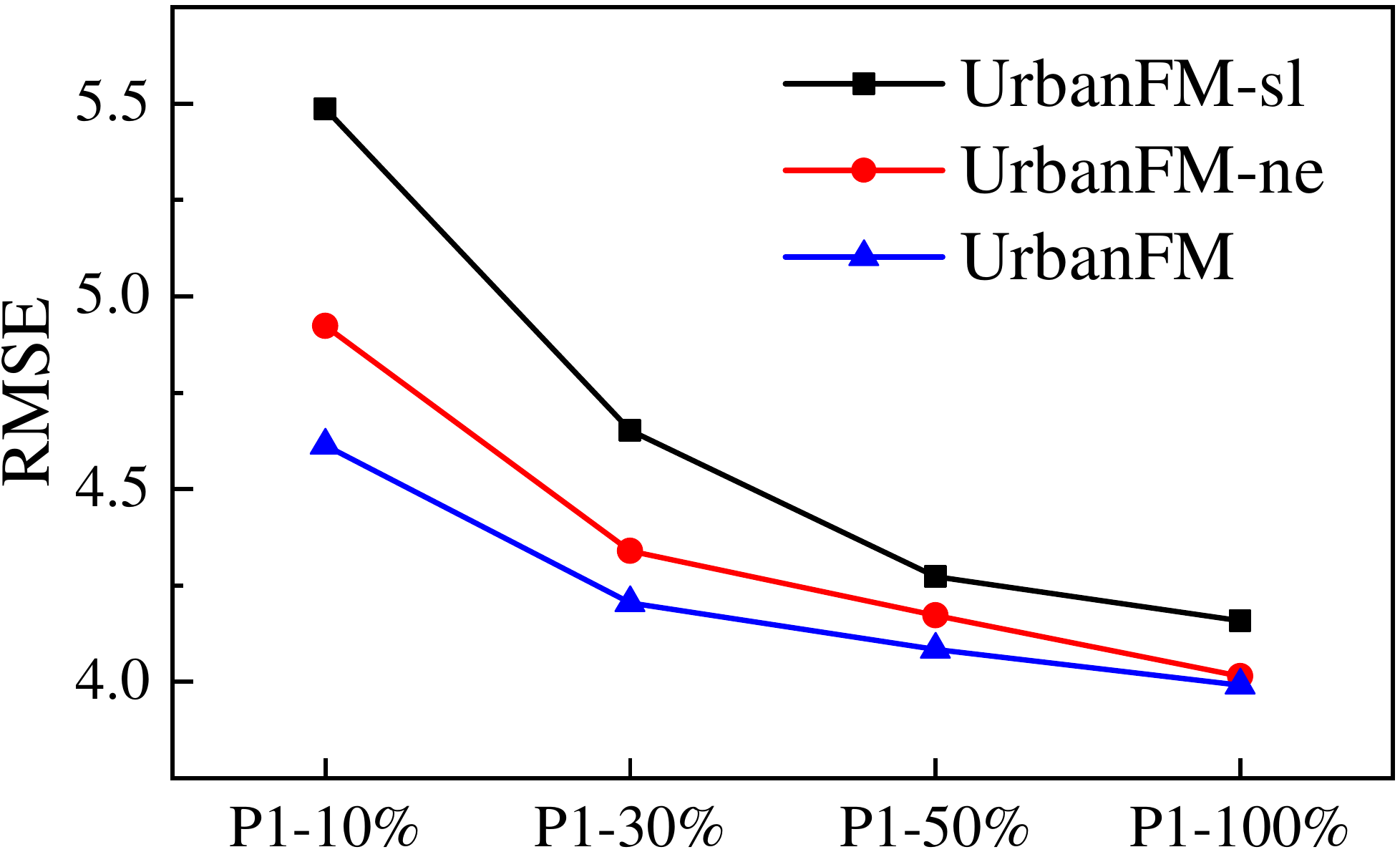}	
		\label{fig:ext_rmse}}
	\end{subfloat}
	\begin{subfloat}[Results on MAE]{
		\includegraphics[width=0.225\textwidth]{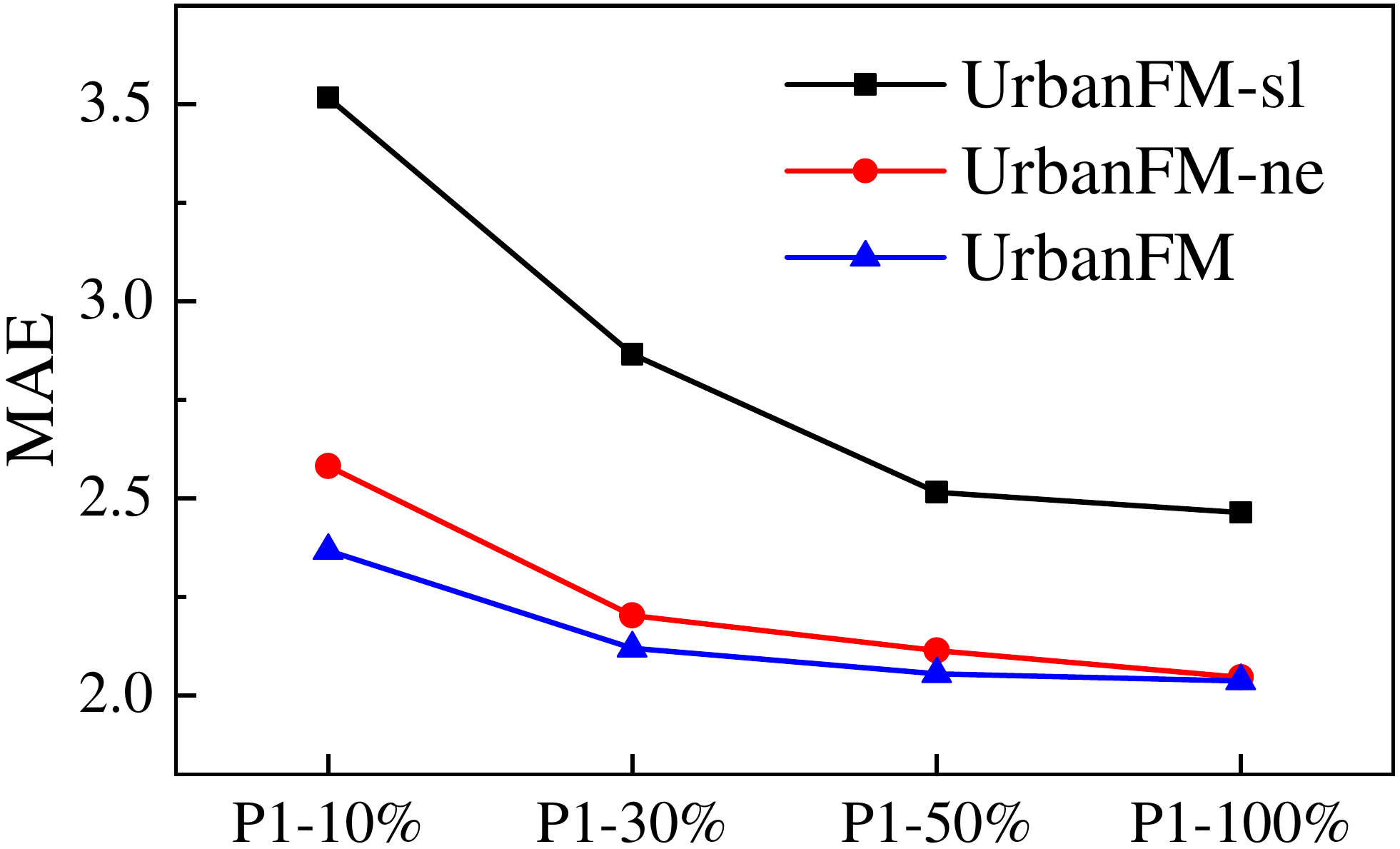}
		\label{fig:ext_mae}}
	\end{subfloat}
	\caption{\textbf{Study on External Factor Fusion}. We reveals the contribution of the fusion network by varying the amount of available training data.}
	\label{fig:ext_effect}
\end{figure}

As shown in Figure~\ref{fig:ext_effect}, the \textit{gap} between UrbanFM and UrbanFM-ne becomes larger as we reduce the number of training data, indicating that external factor fusion plays a more important role in providing a priori knowledge. When the training size grows, the weight for the priori knowledge decreases, as there exists overlaying information between observed traffic flows and external factors. Thus, the network may learn to capture some external impacts when given enough data. Moreover, this trend also occurs between UrbanFM and UrbanFM-sl, which illustrates that the $N^2$-\textit{Normalization} layer provides a strong structural prior to facilitate network training.

%

%
%

\begin{table}[b!]
	\centering
	\renewcommand{\arraystretch}{1.1}

    \begin{subfloat}[Various $M$-$F$ configurations of UrbanFM at $4\times$]{
        \centering
        \label{tab:param_study:1}
		\begin{tabular}{c|lc|rrr}
		\shline
    	\multicolumn{1}{c|}{Methods} & \multicolumn{1}{l}{Settings} & \multicolumn{1}{l|}{\#Params} & \multicolumn{1}{l}{RMSE} & \multicolumn{1}{l}{MAE} & \multicolumn{1}{l}{MAPE} \\
		\hline
		UrbanFM & 16-64 (base) & 1.6M & 4.107 & 2.118 & 0.322 \\
		\hline
	    SRResNet & 20-64 & 1.8M  & 4.317 & 2.586 & 0.725 \\
   		UrbanFM & 20-64 & 1.9M  & 4.094 & 2.101 & 0.321 \\
    	\hline
	    SRResNet & 16-128 & 6.0M  & 4.301 & 2.588 & 0.740 \\
    	UrbanFM & 16-128 & 6.2M  & 4.069 & 2.092 & 0.316 \\
	    \hline
    	SRResNet & 16-256 & 24.2M & 4.178 & 2.418 & 0.614 \\
	    UrbanFM & 16-256 & 24.4M & 4.068 & 2.087 & 0.316 \\
    	\shline
	    \end{tabular}%
	}
    \end{subfloat}
    \hfill
    
    \begin{subfloat}[Various $F$-$M_2$-$R$ configurations of UrbanPy at $8\times$]{
        \centering
        \label{tab:param_study:2}
        \begin{tabular}{c|lc|rrr}
        \shline
        Method & Settings & \#Params & RMSE & MAE & MAPE \\
        \hline
        \multirow{7}{*}{UrbanPy} & 4-64-4 (base) & 4.7M  & 4.572 & 2.237 & 0.352 \\
    \cline{2-6}         & 4-32-4 & 3.0M  & 4.731 & 2.326 & 0.379 \\
              & 4-128-4 & 11.6M & 4.547 & 2.222 & 0.348 \\
    \cline{2-6}         & 2-64-4 & 4.3M  & 4.619 & 2.265 & 0.359  \\
              & 8-64-4 & 5.6M  & 4.581 & 2.240 & 0.351 \\
    \cline{2-6}         & 4-64-1 & 3.9M  & 4.576 & 2.239 & 0.350 \\
              & 4-64-8 & 5.8M  & 4.571 & 2.233 & 0.349 \\
        \shline
        \end{tabular}
    }
     \end{subfloat}
     \caption{\textbf{Study on Configurations}. For both UrbanFM and UrbanPy, we are interested in the key configurations that control the network capability and size of parameters are the filter size $F$, number of residual blocks $M$ and $M_2$, and the number of layers $R$ of the proposal network.}
     \label{tab:param_study}
\end{table}

\begin{figure}[b!]
	\centering
	\includegraphics[width=0.48\textwidth]{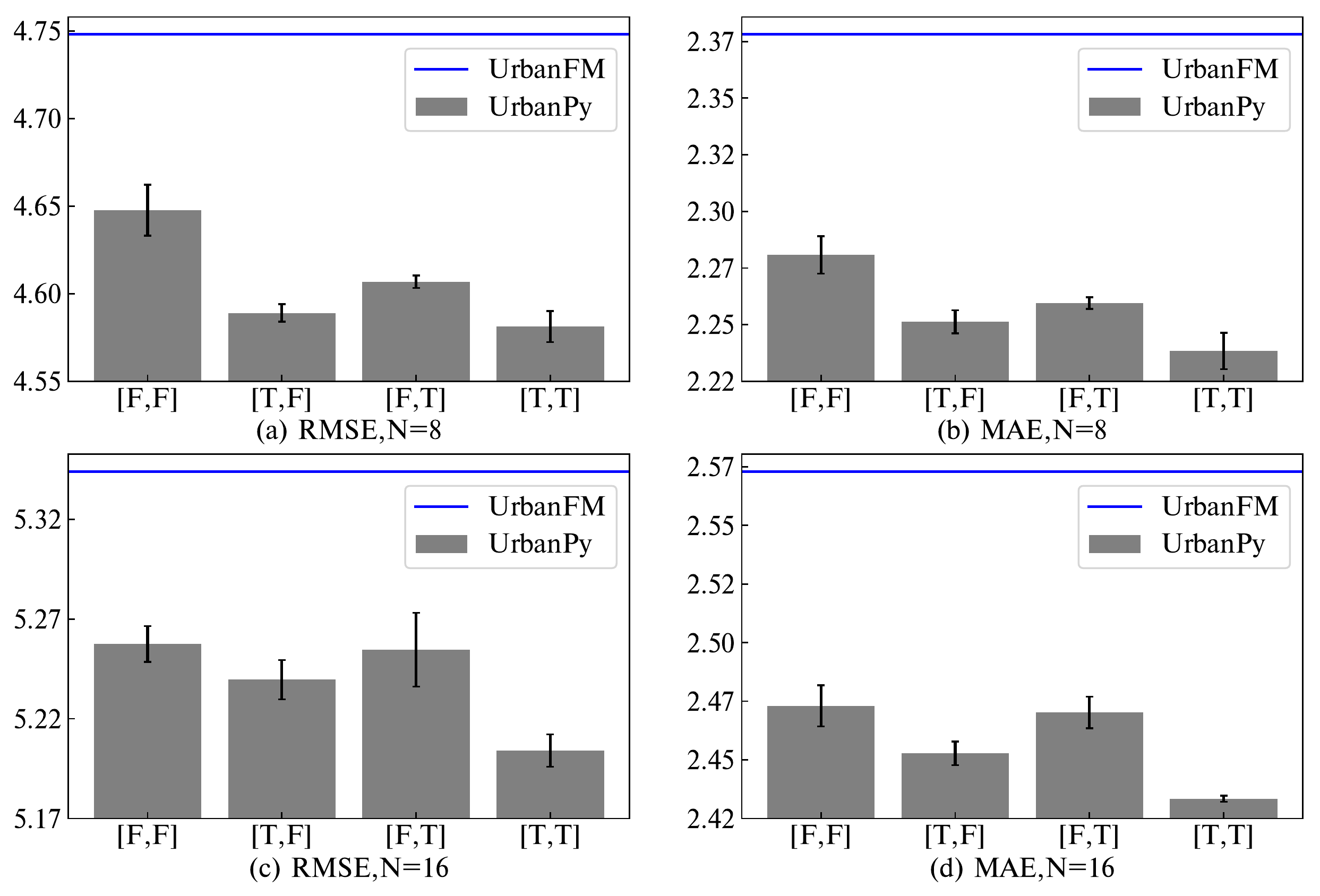}
	\vspace{-1em}
	\caption{\textbf{Ablation study on UrbanPy}. We conduct three experiments on N=8 and N=16 tasks and report the mean and std of RMSE and MAE on the test sets. The configurations involve [Local Structure, Distributional Loss]. We use \emph{T} to denote the present of an element or \emph{F} otherwise.}
	\label{fig:ablation}
\end{figure}
\vspace{1em}
\noindent\textbf{Ablation study on UrbanPy}

\noindent UrbanPy embraces three enhancements over UrbanFM. To reveal the individual contribution of each element, we conduct ablation studies and diagram the results in Figure~\ref{fig:ablation}. It can be seen that UrbanPy with the pyramid architecture alone (i.e., [F,F]) has already outperformed UrbanFM by a large margin in all four settings. This result underscores the importance of the progressive structure and demonstrate its leading role of the improvement of the inference task. By comparing different variants of UrbanPy, it can be seen that the local structure contributes (i.e., [T,F]) a bit more than the distributional loss (i.e., [F,T]). It can also provides more robustness as is witnessed in Figure~\ref{fig:ablation}c. This is not unexpected as the model needs to find a balance between RMSE and the KL-divergence and thus can variate in some situations. Nevertheless, the combination of both elements (i.e., [T,T]) achieves the best performance and good stableness in all settings.

\vspace{1mm}
\noindent\textbf{Study on Configurations}

\noindent Table~\ref{tab:param_study:1} compares the average performance over P1 to P4. Across all hyperparameter settings, UrbanFM consistently outperforms SRResNet, advancing by at least 2.6\%, 13.7\%, and 48.6\%. Besides, this experiment reveals that adding more ResBlocks (larger $M$) or increasing the number of filters (larger $F$) can improve the model performance. However, these also increase the training time and memory space. Considering the tradeoff between training cost and performance, we use 16-64 for UrbanFM as default.

In Table~\ref{tab:param_study:2}, we compare different $M_2$-$F$-$R$ combinations with $4$-$64$-$4$ being the base setting. We observe the following. 1) By increasing $F$ we can see that the network also improves its performance, while the network parameter also blows up quickly. As $F$=128 only outdoes $F$=$64$ marginally, we use the latter setting as default. 2) Different from the effect of $F$, increasing $M_2$ can temper the network performance. This is unlikely due to overfitting as the parameter size remains less than that of $4$-$128$-$4$. Instead, the network can introduce too much zero inputs at the high-level layers as the receptive field at the middle level can already cover the whole coarse-grained input area when $M_2$ is too large. Therefore, We find $4$ a reasonable depth for the residual block. 3) The proposal network is not sensitive to the change of $R$ as is shown in the last two rows. We use $R$=$4$ since we find it is more stable for network training. 

\begin{figure}[t!]
	\centering
	\includegraphics[width=0.49\textwidth]{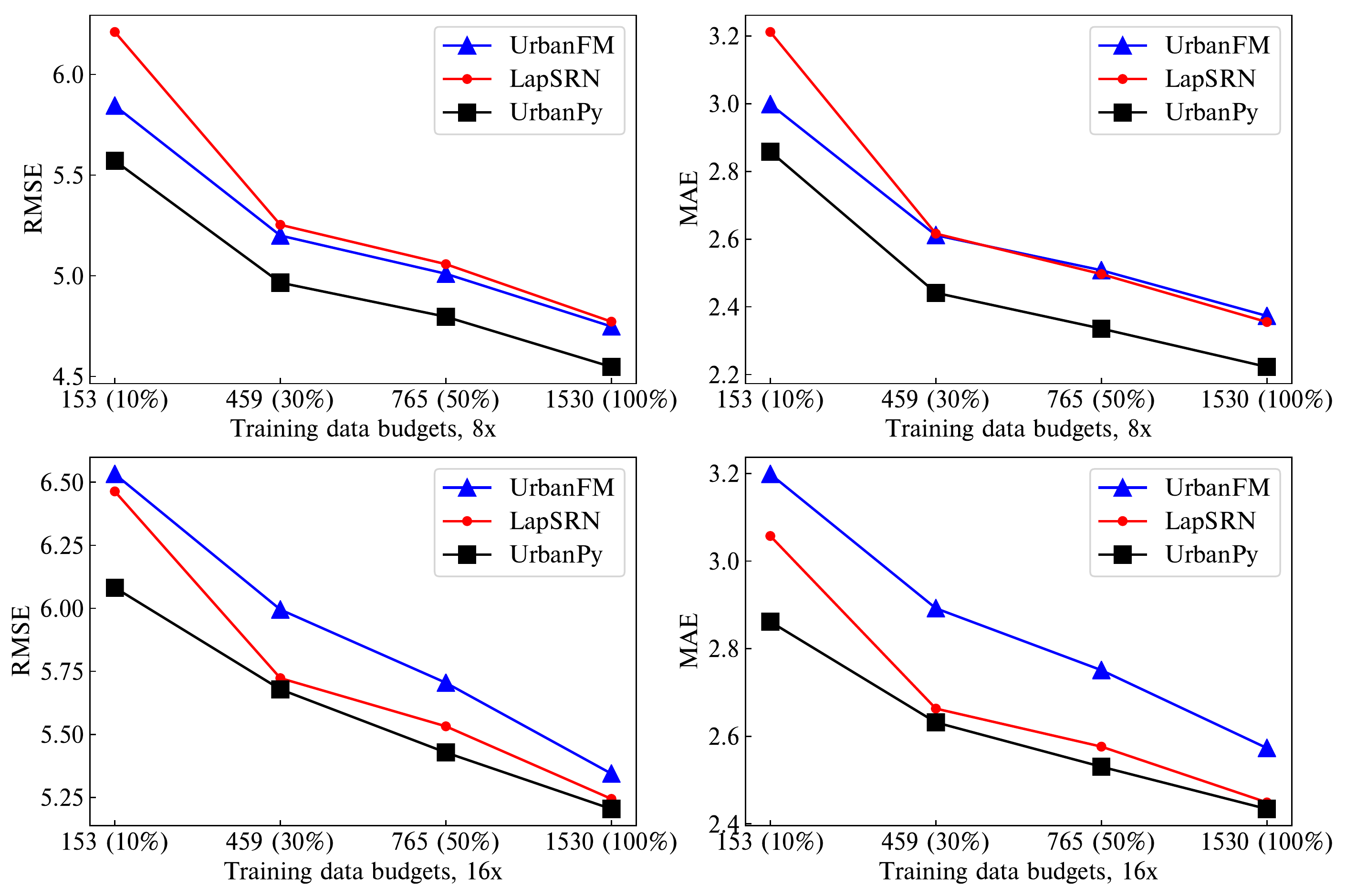}
	\vspace{-2em}
	\caption{\textbf{Robustness along the change of data budget.} We train the models by feeding different amounts of the training data of P1 for large upscaling tasks including $8\times$ and $16\times$.}
	\label{fig:data_eff}
\end{figure}

\vspace{1em}
\noindent\textbf{Study on Robustness to Training Data Budget}

\noindent With the interest of model performance when less training data are available for larger-scale inference tasks, we depict the comparative results with the strongest baseline LapSRN in Figure~\ref{fig:data_eff}. As it illustrates, all models increase their performances as the training data become larger. At $N$=8, UrbanFM remains very competitive and even more data efficient than the LapSRN that enjoys the progressive structure. At $N$=16, LapSRN achieves lower error than UrbanFM as the structure advantages start to overcomes the benefits bought by $N^2$-Normalization and external features when the task becomes too difficult. Therefore, it is no surprise that UrbanPy outperforms the LapSRN in all data budgets for both metrics, as it combines the advances of UrbanFm and benefits from progressive upsampling.

\begin{figure}[!h]
	\centering
	\hspace{1mm}
	\begin{subfloat}[Model comparison.]{
		\includegraphics[width=0.22\textwidth]{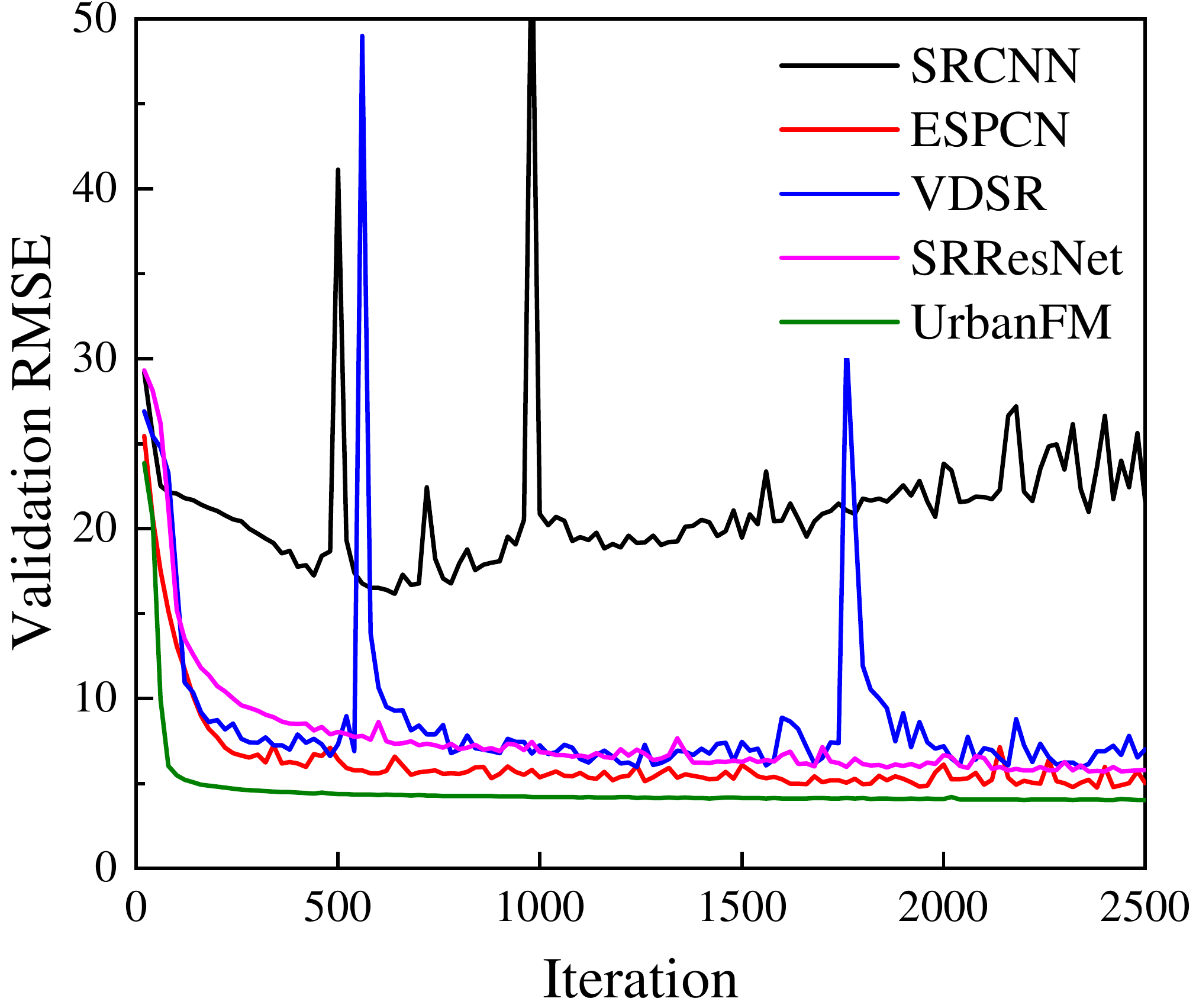}
		\label{fig:effi1}}
	\end{subfloat}
	\begin{subfloat}[Variant comparison.]{
		\includegraphics[width=0.22\textwidth]{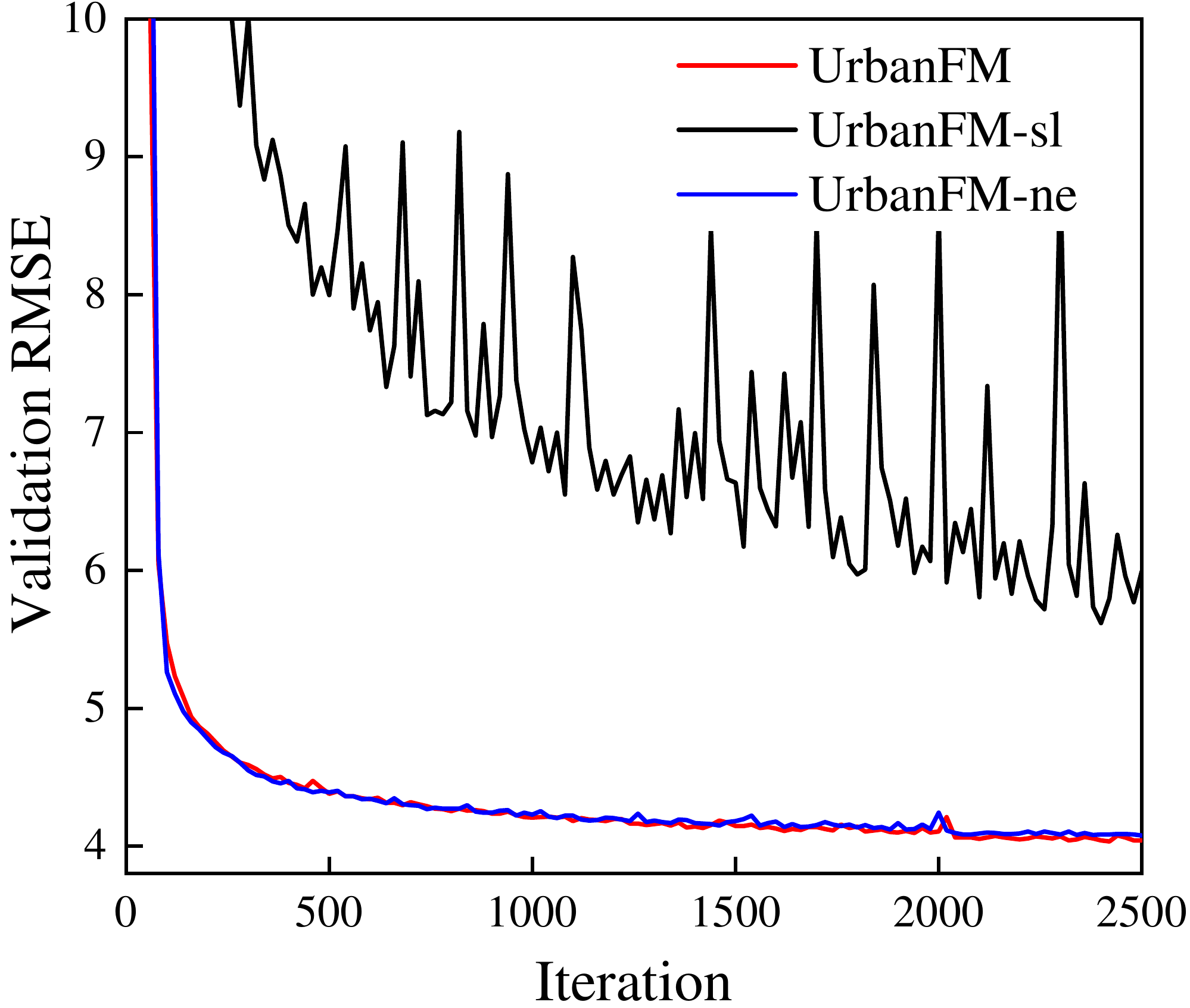}
		\label{fig:effi2}}
	\end{subfloat}
	\begin{subfloat}[Model comparison.]{
		\includegraphics[width=0.217\textwidth]{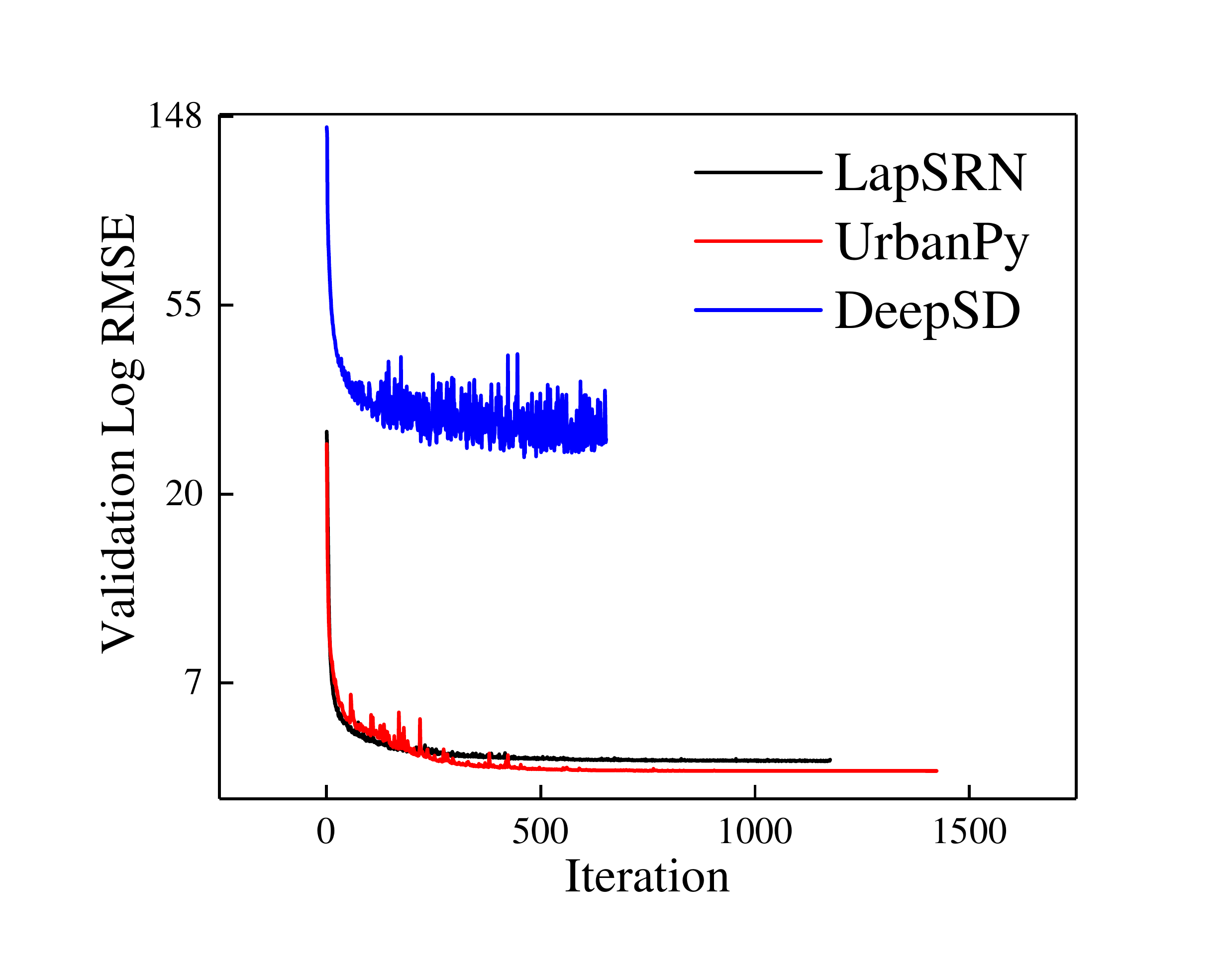}
		\label{fig:effi3}}
	\end{subfloat}
	%
	\hspace{0.1mm}
	\begin{subfloat}[Variant comparison.]{
		\includegraphics[width=0.213\textwidth]{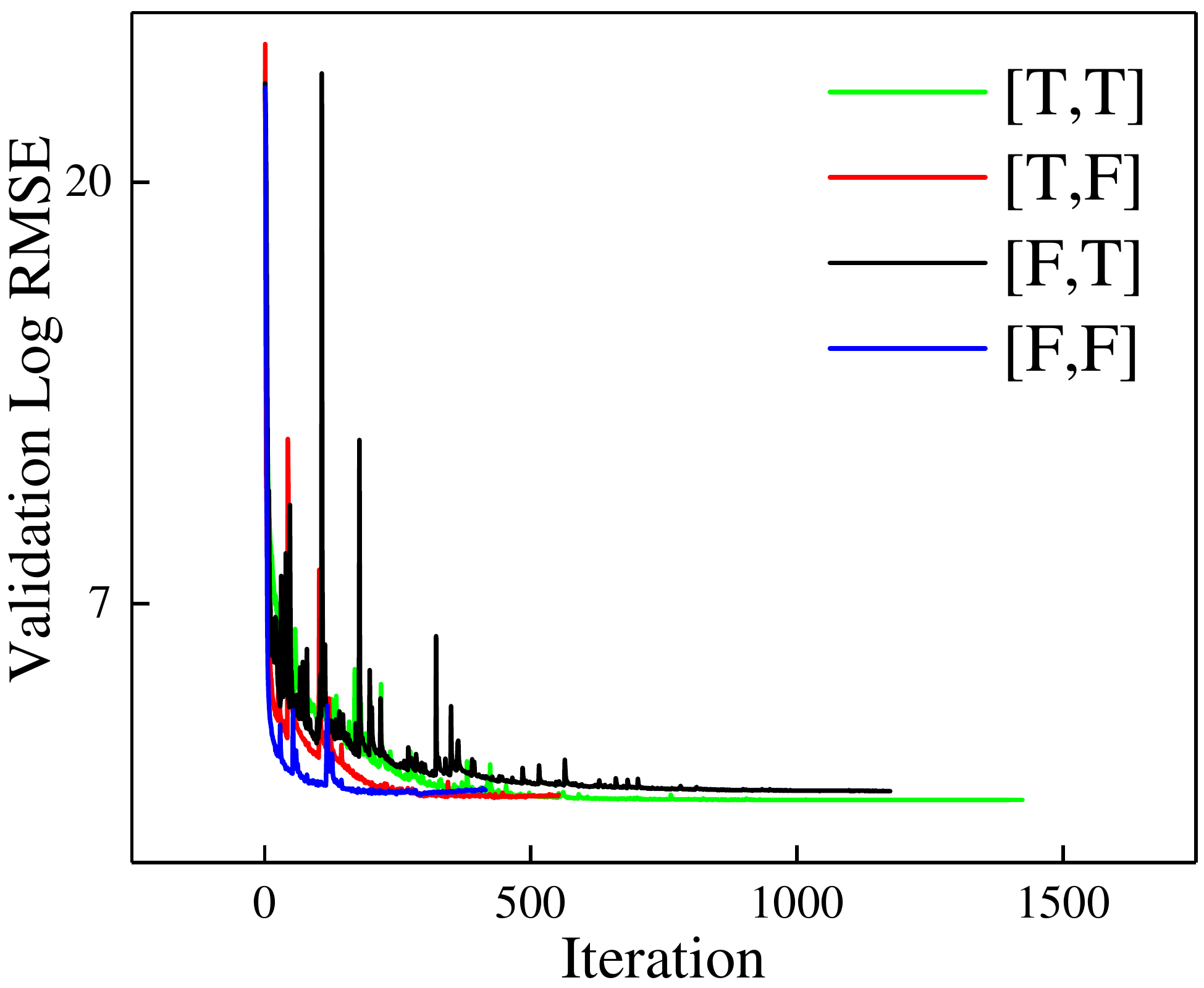}
		\label{fig:effi4}}
	\end{subfloat}
	\caption{\textbf{Convergence speed of various methods.} Figure (a) and (b) show the convergence speed for single-pass methods using 4$\times$ upscaling factors. Differently, we plot logarithm scores in (c) and (d) for progressive methods with $8\times$ for a clearer illustration. Note that we double the training batch size for progressive methods and employ early stopping according to the best validation scores, as their training are generally slower. We use the same notation for variants as in Figure~\ref{fig:ablation}.}
	\label{fig:efficiency}
\end{figure}


\subsubsection{Study on Training Efficiency}

\noindent Figure~\ref{fig:efficiency} plots the RMSE on the validation set during the training phase using P1-100\%. Figure~\ref{fig:efficiency}(a) and \ref{fig:efficiency}(b) delineate that UrbanFM converges much \textit{smoother} and \textit{faster} than the single-pass baselines and its variants. Specifically, \ref{fig:efficiency}(b) suggests such efficiency improvement can be mainly attributed to the $N^2$-\textit{Normalization} layer since UrbanFM-sl converges much slower and fluctuates drastically even it is constrained by $L_s$, when compared with UrbanFM and UrbanFM-ne. This also suggests that learning the spatial correlation is a non-trivial task. Moreover, UrbanFM-ne behaves closely to UrbanFM as external factors fusion affects the training speed subtly when training data are abundant as suggested by the previous experiments.

The convergence curves for progressive methods are depicted by Figure~\ref{fig:efficiency}(c). It illustrates that DeepSD can not converge to the same level as the cascading methods do, due to the inconsistency resides between the stacked components while cascading structure allows training a more coherent network. UrbanPy converges as fast as LapSRN and can be trained continuously longer, as the proposal network is a more powerful component than the simple bilinear interpolation function used by LapSRN. \ref{fig:efficiency}(d) gives a more detailed plot that focuses on UrbanPy and its variants. It can be seen that the [F,F] setting converges smoother then others, however, stops earlier and fails to improve further. [F,T] shows large fluctuation during training as the KL-divergence can reduce the stableness. Nevertheless, [T,T] shows both smooth convergence curve and continuous improvement, which explains why combining both local structure and distributional loss can outperform the state-of-the-art methods. 


\subsubsection{Visualization}
\begin{figure}[!t]
	\centering
	\includegraphics[width=0.475\textwidth]{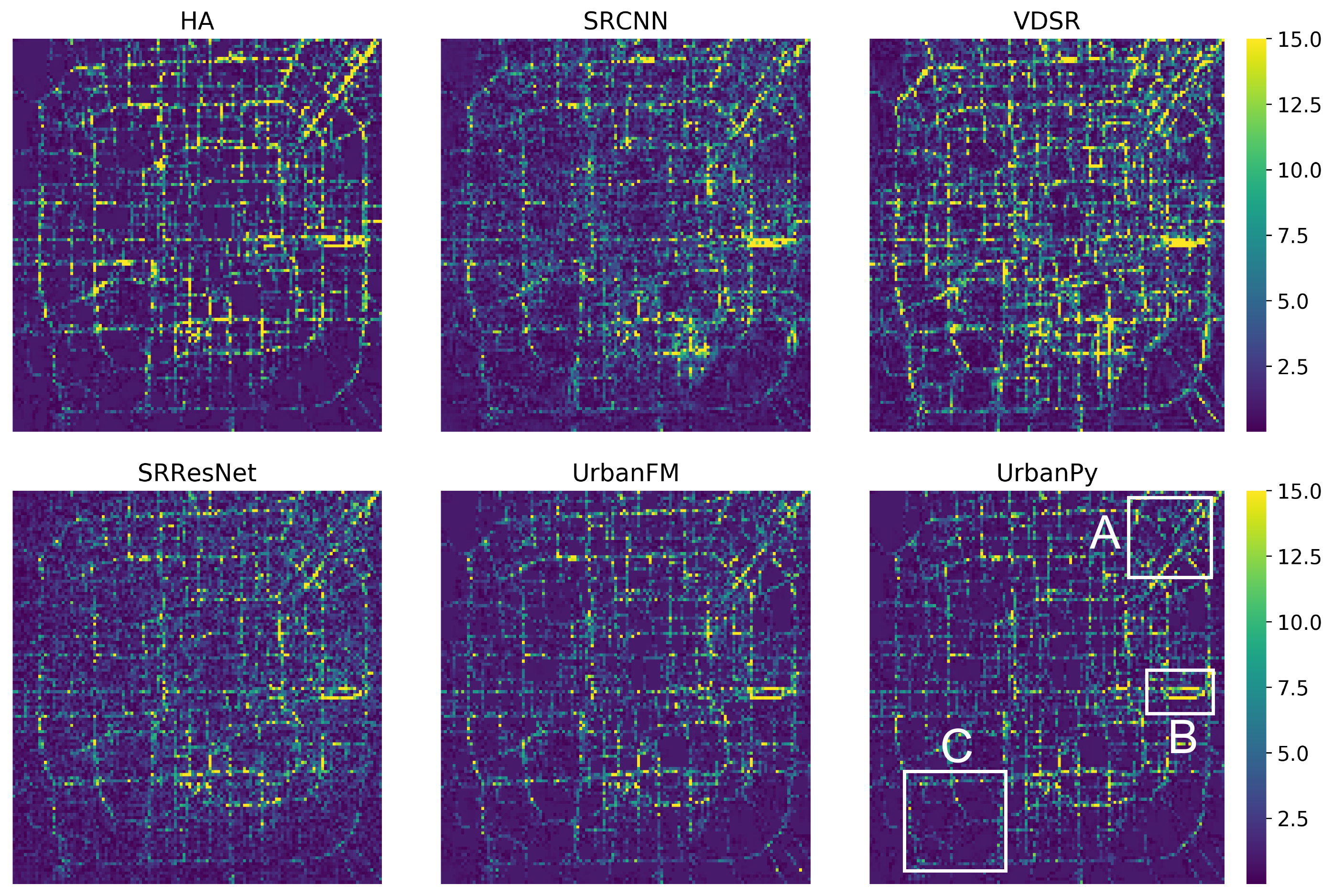}
	\vspace{-1em}
	\caption{\label{fig:diff} Visualization for inference errors among different methods. Best view in color.}	
	\vspace{-1em}
\end{figure}

\noindent1)~\textit{Inference error.} Figure~\ref{fig:diff} displays the inference error $\lVert \mathbf{X}^f-\tilde{\mathbf{X}}^f \rVert_{1,1}$ from our methods and the other four baselines for a sample at the 4$\times$ task, where a brighter pixel indicates a larger error. Contrast with the baseline methods, both UrbanFM and UrbanPy achieves higher fidelity for totality and in detail, which corresponds to the quantitive results from Table~\ref{tab:table1}. For instance, areas A and B are "hard areas" to be inferred, as A (Sanyuan bridge, the main entrance to downtown) and B (Sihui bridge, a huge flyover) are two of the top congestion points in Beijing. Traffic flow of these locations usually fluctuates drastically and quickly, resulting in higher inference errors. Nonetheless, Our methods remain to produce better performances in these areas. Another observation is that the SR methods (SRCNN, VDSR, and SRResNet) tend to generate blurry images as compared to structural methods (HA and our methods). For instance, even if there is zero flow in area C, SR methods still generate error pixels as they overlap the predicted patches. This suggests the FUFI problem does differ from the ordinary SR problem and requires specific designs.

\begin{figure}[!t]
	\centering
	\includegraphics[width=0.46\textwidth]{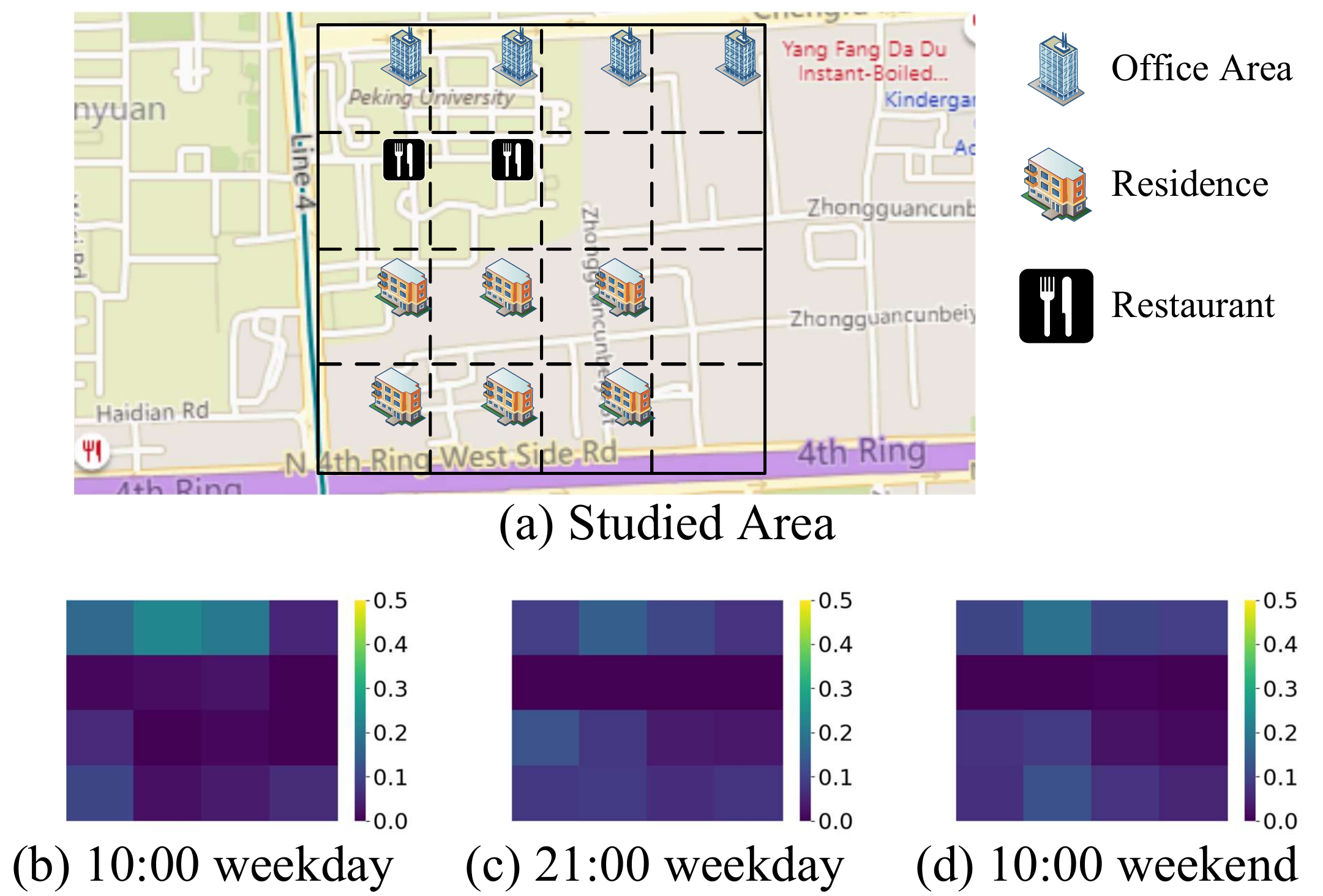}
	\vspace{-1em}
	\caption{\label{fig:case_study} Case study on a superregion near Peking University. See our github for further dynamic analysis on this area.}	
	\vspace{-1em}
\end{figure}

\vspace{1em}
\noindent2)~\textit{External influence.} Figure~\ref{fig:case_study}(a)-(d) portray that the \textit{inferred} distribution over subregions varies along with external factor changes. To stay succinct, we present the results of UrbanFM only as UrbanPy produces similar visualization regarding external factors. On weekdays, at 10 a.m., people had already flowed to the office area to start their work (b); at 9 p.m., many people had returned home after a hard-working day (c). On weekends, most people stayed home at 10 a.m. but some industrial researchers remained working in the university labs. This result proves that our methods indeed capture the external influence and learns to adjust the inference accordingly.

\subsubsection{Results on HappyValley}
Table~\ref{tab:hv} shows model performances using the HappyValley dataset. Note that in this experiment, we do not include DeepSD, since this task contains only $2\times$ upscaling such that DeepSD degrades to SRCNN in this case. One important trait of the HappyValley dataset is that it contains more spikes on the fine-grained flow distribution, which results in a much larger RMSE score versus that in the TaxiBJ task. Nonetheless, in the \textit{single-pass} branch UrbanFM remains the winner method outperforming the best baseline by 3.5\%, 7.8\%, and 22\%; the UrbanFM-ne still holds the runner-up position. Moreover, it is not unexpected to see LapSRN is worse than UrbanFM as the former shows no progressive superiority over UrbanFM in this task. Move on to the \textit{progressive} branch. Though the [F,F] variants show worse performances than UrbanFM, as the compositional architecture can complicate the training when the task is as simple as 2$\times$ upscaling, the full models of UrbanPy can provide better scores then its single-pass counterpart, which validates the usefulness of the two components even when no structural advance can be exploited.   

To summarize, this collection of experiment prove that our methods not only work on the large-scale scenario, but is also adaptable to smaller areas, which concludes our empirical studies.
\begin{table}[tbp]
  \centering
  \renewcommand{\arraystretch}{1.1}

  \caption{\textbf{Results comparison on Happy Valley.} We evaluate the task with 2$\times$ upscaling for this area. All models are selected based on the best validation performance and test results are presented.}
    \begin{tabular}{c|cc|ccc}
    \shline
    Methods & Settings & Params & RMSE  & MAE   & MRE \\
    \hline
    MEAN & x     & x     & 9.206 & 2.269 & 0.799 \\
    HA & x     & x     & 8.379 & 1.811 & 0.549 \\
    SRCNN & 768   & 7.4M  & 8.291 & 2.175 & 0.816 \\
    ESPCN & 768   & 7.4M  & 8.156 & 2.155 & 0.805 \\
    VDSR & 16-64 & 0.6M  & 8.490 & 2.128 & 0.756 \\
    SRResNet & 16-128 & 5.5M  & 8.318 & 1.941 & 0.679 \\
    UrbanFM-sl & 16-128 & 5.5M  & 8.312 & 1.939 & 0.677 \\
    UrbanFM-ne & 16-128 & 5.5M  & 8.138 & 1.816 & 0.537 \\
    UrbanFM & 16-128 & 5.6M & \textbf{8.030} & \textbf{1.790} & \textbf{0.531} \\
    \hline
    LapSRN & 10-128 & 3.2M  & 8.249 & 1.832 & 0.547 \\
    UrbanPy-FF & 8-64-4 & 1.2M  & 8.280 & 1.879 & 0.587 \\
    UrbanPy-TT & 8-64-4 & 1.3M & \textbf{8.028} & 1.749 & 0.523\\
    UrbanPy-FF & 8-128-4 & 4.4M  & 8.184 & 1.900 & 0.618 \\
    UrbanPy-TT & 8-128-4 & 4.4M  & 8.332 & \textbf{1.732} & \textbf{0.508} \\
    \shline
    \end{tabular}%
  \label{tab:hv}%
  \vspace{-1em}
\end{table}%

\vspace{1em}
\noindent\textbf{Limitation}

\noindent While our methods demonstrates leading performance for both low-scale (UrbanFM) and large-scale (UrbanPy) urban flow inference tasks, the current structure accepts the regular partition of the urban area. For non-regular partition, we need to use a graph to represent the locations as nodes and connections between locations (e.g., road networks) as the edges. Besides, the UrbanPy learns slower than its single-pass counterpart (typically 4 times slower as can be seen in Figure~\ref{fig:efficiency}) as the dynamics become more complicated with the pyramid structure, which is also noted by ~\cite{lai2018fast,wang2018fully}. Nevertheless, this is a trade-off between training efficiency and inference performance. We suggest that when the required upsampling scale is large, the UrbanPy is a more favorable choice; if the training time is of the key concern or the scale is small, we should opt for the UrbanFM model.

\section{Related Work}\label{sec:relatedwork}

\subsection{Image Super-Resolution}\label{sec:SISR}
Single image super-resolution (SISR), which aims to recover a high-resolution (HR) image from a single low-resolution (LR) image, has gained increasing research attention for decades. This task finds direct applications in many areas such as face recognition \cite{gunturk2003eigenface}, fine-grained crowdsourcing \cite{thornton2006sub} and HDTV \cite{park2003super}. Over the years, the computer vision community has presented many efforts in developing SISR algorithms that can be largely categorized into two: single-pass and progressive methods.  

\subsubsection{Single-pass methods}

Single-pass methods process coarse-grained images in one or multiple consecutive upsampling steps. Early upsampling techniques exploited interpolation methods such as bicubic interpolation and Lanczos resampling \cite{duchon1979lanczos}. Also, several studies utilized statistical image priors \cite{sun2008image, tai2010super} to achieve better performances. Advanced works aimed at learning the non-linear mapping between LR and HR images with neighbor embedding \cite{chang2004super} and sparse coding \cite{yang2010image,timofte2014a}. However, these approaches are still inadequate to reconstruct realistic and fine-grained textures of images.

Recently, a series of models based on deep learning has achieved great success in terms of SISR as they do not require any human-engineered features and show the state-of-the-art performance. Since \citeauthor{dong2016srcnn} \cite{dong2016srcnn} first proposed an end-to-end mapping method represented as CNNs between the low-resolution (LR) and high-resolution (HR) images, various CNN based architectures have been studied for SR. Among them, \citeauthor{shi2016espcn} \cite{shi2016espcn} introduced an efficient sub-pixel convolutional layer which is capable of recovering HR images with very little additional computational cost compared with the deconvolutional layer at training phase. Inspired by VGG-net for ImageNet classification \cite{simonyan2014vgg}, a very deep CNN was applied for SISR in \cite{kim2016vdsr}. However, training a very deep network for SR is really hard due to the small convergence rate. \citeauthor{kim2016vdsr} \cite{kim2016vdsr} showed residual learning speed up their training phase and verified that increasing the network depth could contribute to a significant improvement in SR accuracy. 

The general process of SISR methods (i.e., feature extraction followed by SR image recovery) inspires our solution for FUFI. However, these approaches are not suitable for the FUFI problem since the flow data present a very specific hierarchical structure with regard to natural images, as such, the related arts cannot be simply applied to our application in terms of efficiency and effectiveness. 

\subsubsection{Progressive methods}
Though single-pass methods demonstrate useful performances at small-scale upsampling (typically 2$\times$ and $4\times$), these methods encounter difficulties when dealing with large-scale super-resolution tasks (e.g., 8$\times$)~\cite{lai2017deep}. This can be attributed to the abrupt upsampling based on low-level features and utilize only one supervision signal at the output end. To tackle this problem, several nascent works~\cite{lai2017deep,lai2018fast,wang2018fully} proposed progressive models based on laplacian pyramid, where the network aimed to learn the upsampled residuals and perform upsampling by aggregating the residuals with interpolated images. This inspires the cascading design of our UrbanPy architecture.

Apart from super-resolving classical images, there are limited studies that utilized super-resolution methods to solve real-world problems in the urban area. In particular, two very recent works \cite{vandal2017deepsd,zong2019deepdpm} employ the same strategy of stacking SRCNN~\cite{dong2016srcnn} for  two different tasks: ~\citeauthor{vandal2017deepsd}~\cite{vandal2017deepsd} aimed at statistical downscaling of climate and earth system simulations based on observational and topographical data; likewise, ~\citeauthor{zong2019deepdpm}~\cite{zong2019deepdpm} addressed the task of inferring fine-grained population density by treating the population heat maps as images.

Different from the related arts that directly target the modeling on pixel values, we instead model the distributions over the superregions and their fine-grained counterparts, by doing which we are able to capture the essence of the FUFI problem. Moreover, we also include the external features which are very unique in the urban scenario.

\subsection{Urban Flows Analysis}
Due to the wide applications of traffic analysis and the increasing demand for real-time public safety monitoring, urban flow analysis has recently attracted the attention of a large amount of researchers \cite{zheng2014urban}. \citeauthor{zheng2014urban} \cite{zheng2014urban} first transformed public traffic trajectories into other data formats, such as graphs and tensors, to which more data mining and machine learning techniques can be applied. Based on our observation, there were several previous works \cite{song2014prediction,fan2015citymomentum} forecasting millions, or even billions of individual mobility traces rather than aggregated flows in a region.

Recently, researchers have started to focus on city-scale traffic flow prediction \cite{hoang2016fccf}. Inspired by deep learning techniques that power many applications in modern society \cite{lecun2015deep}, a novel deep neural network was developed by \citeauthor{zhang2016dnn} \cite{zhang2016dnn} to simultaneously model spatial dependencies (both near and distant), and temporal dynamics of various scales (i.e., closeness, period and trend) for citywide crowd flow prediction. Following this work, \citeauthor{zhang2017deep} \cite{zhang2017deep} further proposed a deep spatio-temporal residual network to collectively predict inflow and outflow of crowds in every city grid. Apart from the above applications, very recently \citeauthor{liang2019urbanfm}~\cite{liang2019urbanfm} presented UrbanFM, the first work to the best of our knowledge to solve the novel FUFI problem in urban scenario. In this paper we further extend the capability of UrbanFM to solve larger-scale inference tasks by presenting the UrbanPy framework.

\section{Conclusion}\label{sec:conclusion}
In this paper, we have formalized the fine-grained urban flow inference problem and two versions of deep neural network-based methods to solve it. The preliminary version (i.e., UrbanFM) focuses on addressing the two specific challenges of the problem through embedding the hierarchical structure in the model and generating a comprehensive representation for external factors. Build upon the key components of UrbanFM, we present a more advanced version named UrbanPy by employing the progressive upsampling strategy, which resolves the defects of UrbanFM when tackling larger-scale inference tasks. We have conducted extensive experiments, both qualitatively and quantitively to study the actual performance of the models using the TaxiBJ dataset and HappyValley datasets. The empirical studies and visualizations have supported the advantages of both UrbanFM and UrbanPy on both efficiency and effectiveness. Codes are also published for the community~\footnote{https://github.com/ouyangksoc/UrbanPy}.

We have also discussed the limitation of the current work, which is mainly due to the learning dynamic of the pyramid structure and remains an open problem. For our future work, we are interested in improving the learning efficiency of the UrbanPy framework, by curriculum strategy~\cite{wang2018fully} or exploring differnet network structures~\cite{zhang2018image}.


%


\ifCLASSOPTIONcaptionsoff
  \newpage
\fi
\bibliographystyle{IEEEtranN}
\bibliography{references}

\vspace{-5em}
\begin{IEEEbiography}
    [{\includegraphics[width=1in,height=1.25in,clip,keepaspectratio]{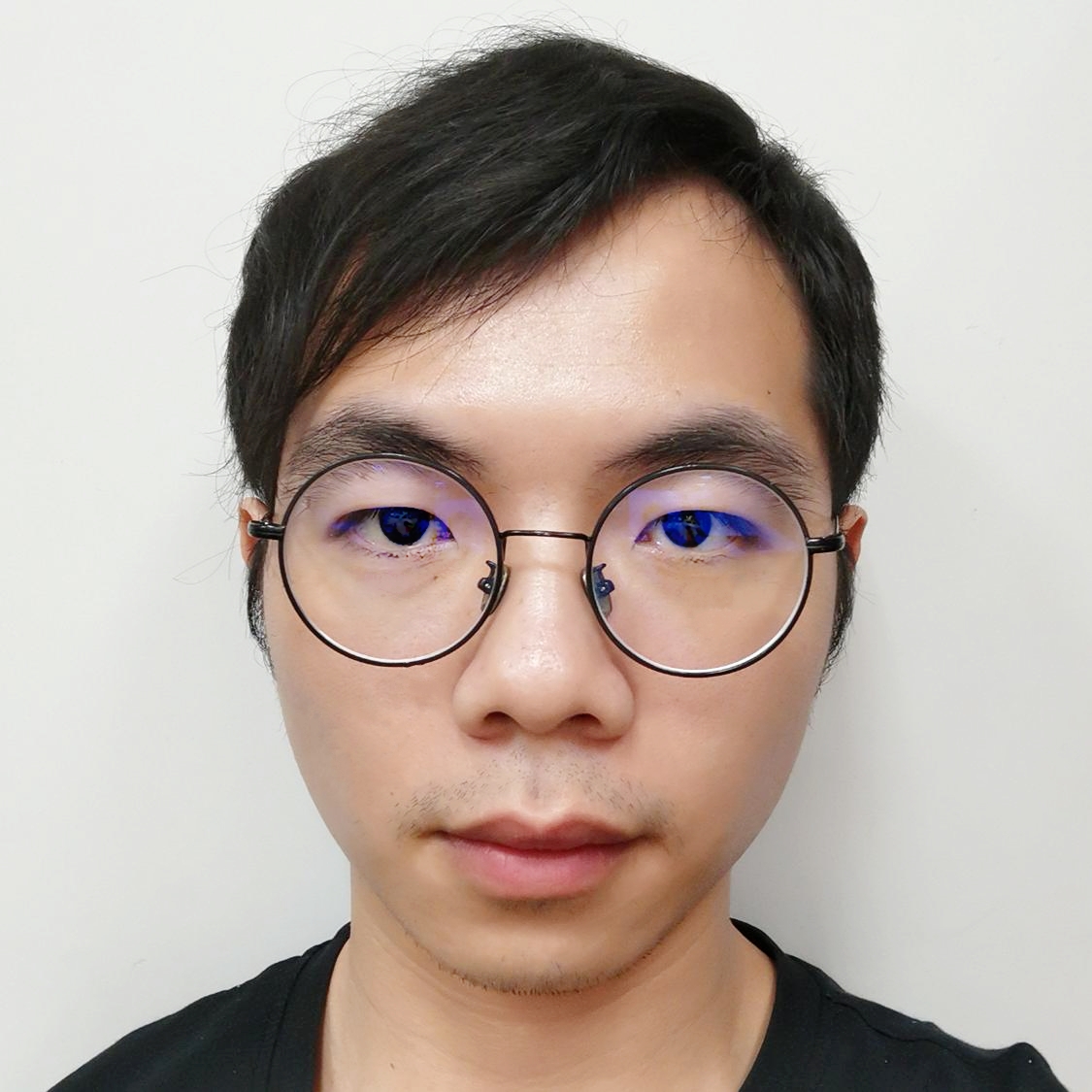}}]{Kun Ouyang} is currently a PhD candidate in Computer Science at the National University of Singapore (NUS). He is also a research scholar in SAP Singapore. He obtained his B.E. degree in the Internet of Things, in Wuhan University in 2015. His research interests include human mobility analytics, spatial-temporal data mining and deep learning.
\end{IEEEbiography}

\vspace{-5em}
\begin{IEEEbiography}
[{\includegraphics[width=1in,height=1.25in,clip,keepaspectratio]{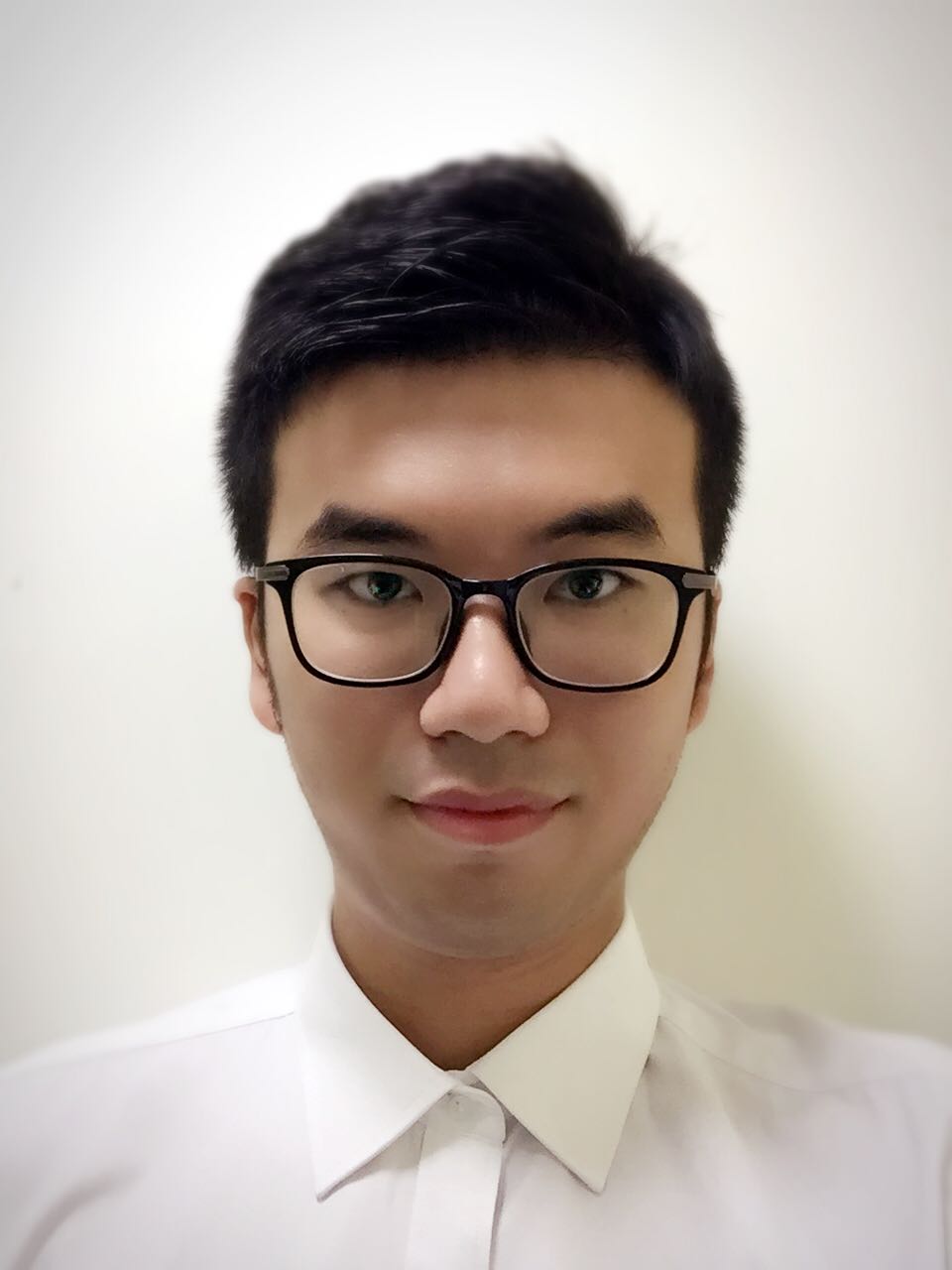}}]{Yuxuan Liang}
 is currently pursuing his Ph. D. degree at School of Computing, National University of Singapore. His research interests mainly lie in machine learning, deep learning and their applications in urban areas.
\end{IEEEbiography}

\vspace{-5em}
\begin{IEEEbiography}
    [{\includegraphics[width=1in,height=1.25in,clip,keepaspectratio]{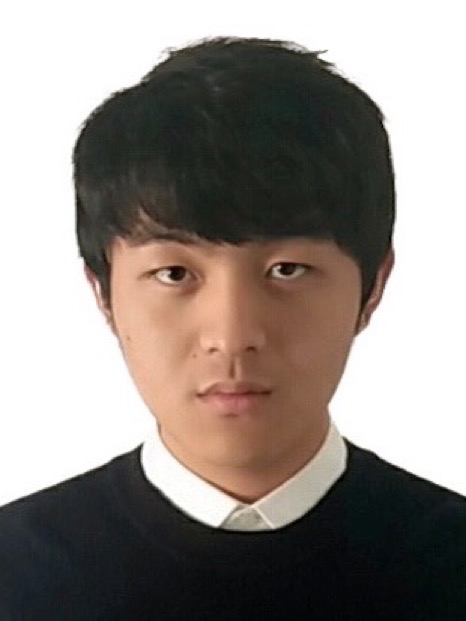}}]{Zekun Tong} received the B.E. degree in computer science and technology from Xidian University, Xi an, China, in 2018.
He is currently pursuing the Ph.D. degree in Industrial System Engineering with the School of Engineering, National University of Singapore, Singapore. His research interests include machine learning, optimization and data analytics.
\end{IEEEbiography}

\vspace{-5em}
\begin{IEEEbiography}
    [{\includegraphics[width=1in,height=1.25in,clip,keepaspectratio]{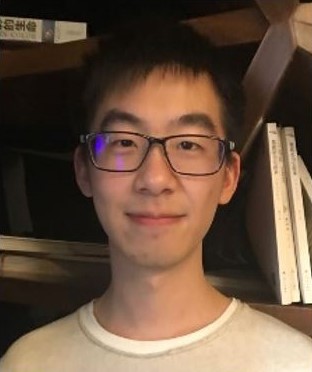}}]{Sijie Ruan} is a Ph.D. student in the School of Computer Science and Technology, Xidian University. He received his B.E. degree from Xidian University in 2017. His research interests include urban computing, spatio-temporal data mining, and distributed systems. He was an intern in MSR Asia from 2016 to 2017. He is now a research intern in JD Intelligent Cities Research, under the supervision of Prof. Yu Zheng and Dr. Jie Bao.
\end{IEEEbiography}

\vspace{-5em}
\begin{IEEEbiography}
    [{\includegraphics[width=1in,height=1.25in,clip,keepaspectratio]{img/bio/ye.pdf}}]{Ye Liu} Dr. Ye Liu received his Ph.D degree from National University of Singapore. Before that he received the M.Sc. degree from Peking University. His research interests lie mainly in the areas of urban computing, ubiquitous computing, knowledge graph, data mining, machine learning, artificial intelligence and their applications in human activity analysis, social network analysis, knowledge graph, urban computing, computational advertising, recommendation.
\end{IEEEbiography}

\vspace{-5em}
\begin{IEEEbiography}[{\includegraphics[width=1in,height=1.25in,clip,keepaspectratio]{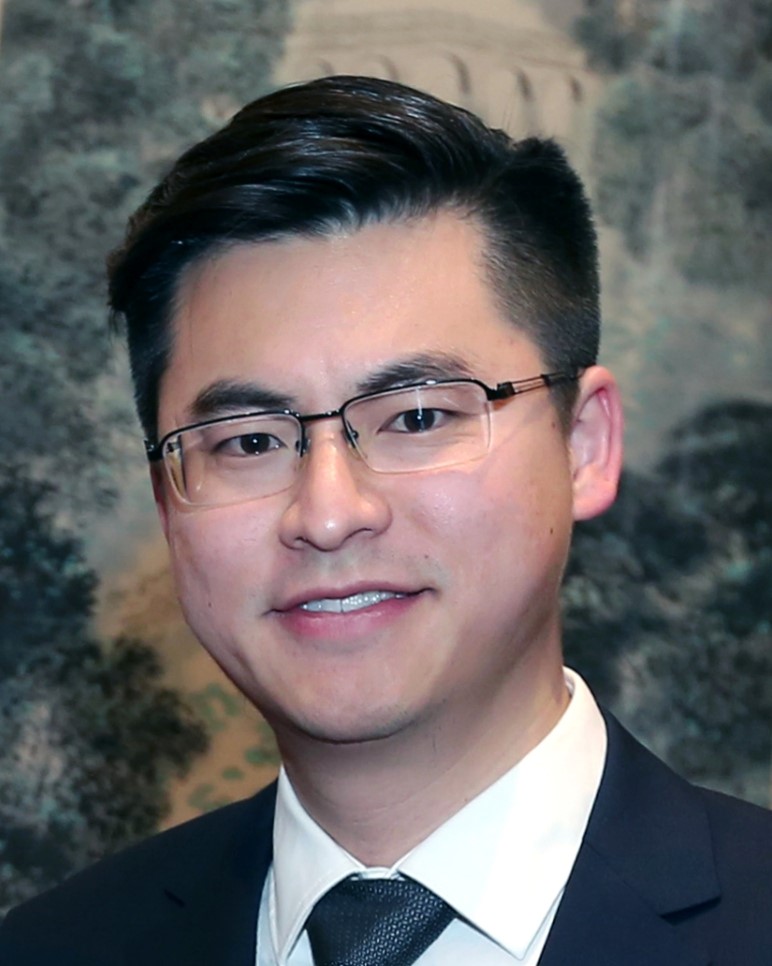}}]{Yu Zheng} is a Vice President of JD.COM and Chief Data Scientist at JD Digits, passionate about using big data and AI technology to tackle urban challenges. His research interests include big data analytic, spatio-temporal data mining, machine learning, and artificial intelligence. He also leads the JD Urban Computing Business Unit as the president and serves as the director of the JD Intelligent City Research. Before joining JD, he was a senior research manager at Microsoft Research. Zheng is also a Chair Professor at Shanghai Jiao Tong University, an Adjunct Professor at Hong Kong University of Science and Technology.
\end{IEEEbiography}

\vspace{-5em}
\begin{IEEEbiography}
    [{\includegraphics[width=1in,height=1.25in,clip,keepaspectratio]{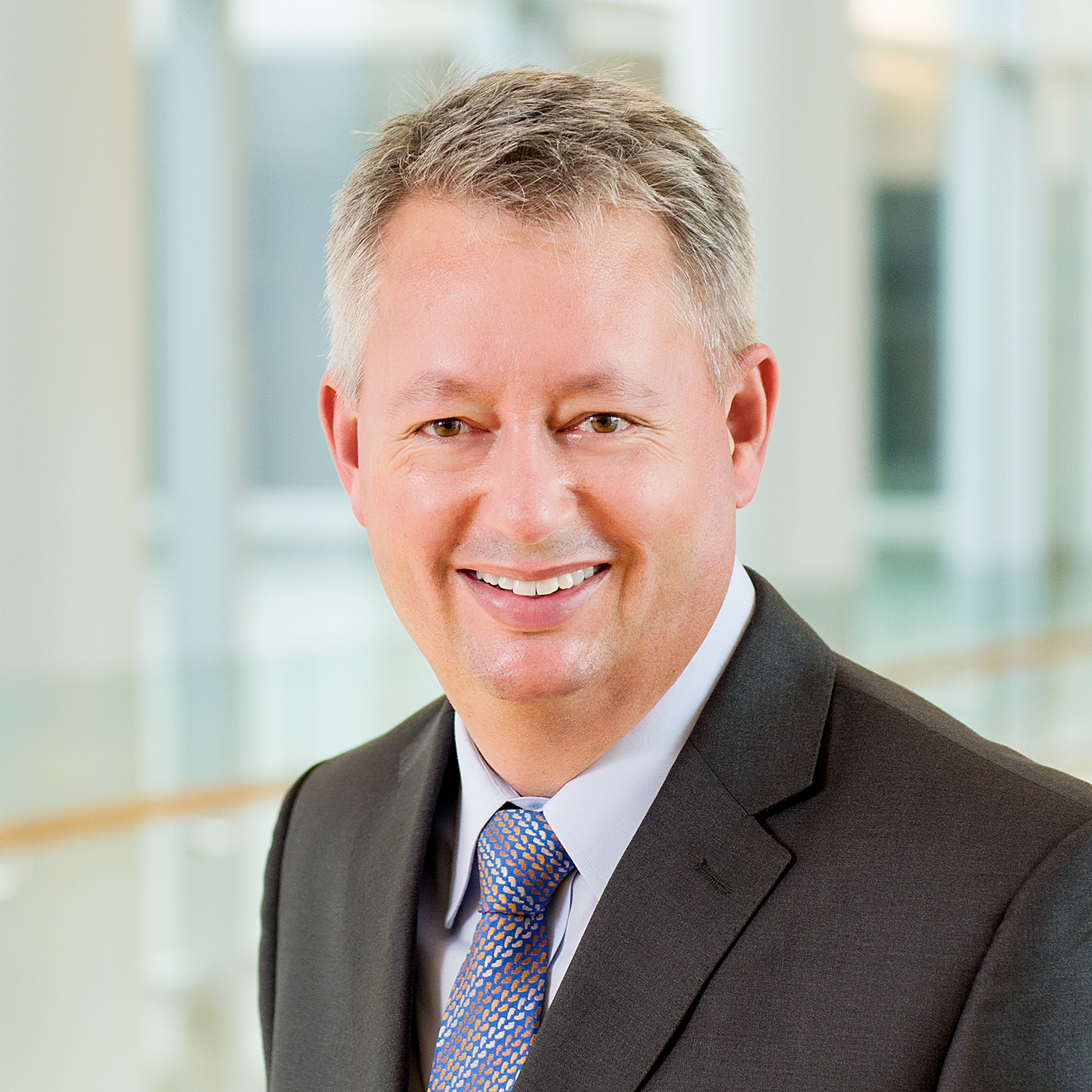}}]{David S. Rosenblum} is Provost's Chair Professor of Computer Science at the National University of Singapore (NUS). His research interests span many problems in software engineering, distributed systems and ubiquitous computing, and his current research focuses on probabilistic verification, uncertainty in software testing, and machine learning.  He is a Fellow of the ACM and IEEE and was previously Editor-in-Chief of the ACM Transactions on Software Engineering and Methodology (ACM TOSEM) and Chair of the ACM Special Interest Group in Software Engineering (ACM SIGSOFT).  He has received two "test-of-time" awards for his research papers, including the ICSE 2002 Most Influential Paper Award for his ICSE 1992 paper on assertion checking, and the inaugural ACM SIGSOFT Impact Paper Award in 2008 for his ESEC/FSE 1997 on Internet-scale event observation and notification (co-authored with Alexander L. Wolf). He also received the ACM SIGSOFT Distinguished Service Award in 2018.
\end{IEEEbiography}

%



%




\end{document}